\documentclass[12pt]{article}
\usepackage{fullpage,graphicx,psfrag,amsmath,amsfonts,verbatim}
\usepackage[small,bf]{caption}\usepackage{amssymb}
\usepackage{enumitem}
\usepackage{tikz}
\usepackage{authblk}
\usepackage{textcomp}
\usepackage{mathtools}
\usepackage{booktabs}
\usepackage{multirow}
\usepackage{subcaption}

\graphicspath{{figures/}}

\usepackage{hyperref}
\usepackage{wasysym}
\hypersetup{ colorlinks   = true,    
urlcolor     = blue,    
linkcolor    = blue,    
citecolor    = red      
}

\newcommand{\reals}{{\mbox{\bf R}}}
\newcommand{\naturals}{{\mbox{\bf N}}}
\newcommand{\integers}{{\mbox{\bf Z}}}

\newcommand{\eg}{{\it e.g.}}
\newcommand{\ie}{{\it i.e.}}

\newcommand{\BEAS}{\begin{eqnarray*}}
\newcommand{\EEAS}{\end{eqnarray*}}
\newcommand{\BEA}{\begin{eqnarray}}
\newcommand{\EEA}{\end{eqnarray}}
\newcommand{\BEQ}{\begin{equation}}
\newcommand{\EEQ}{\end{equation}}
\newcommand{\BIT}{\begin{itemize}}
\newcommand{\EIT}{\end{itemize}}

\newcounter{algorithmctr}
\renewcommand{\thealgorithmctr}{\arabic{algorithmctr}}
   {\mbox{}\\*[\parskip]\begin{minipage}{\linewidth}%
       \refstepcounter{algorithmctr}\begin{list}{}{%
       \setlength{\rightmargin}{0\linewidth}%
       \setlength{\leftmargin}{.05\linewidth}}%
       \rmfamily\small
       \item[]{\setlength{\parskip}{0ex}\hrulefill\par%
        \nopagebreak{\bfseries\textsf{Algorithm \thealgorithmctr~}}}}%
   {{\setlength{\parskip}{-1ex}\nopagebreak\par\hrulefill\\*[2ex]\par}%
   \end{list}\end{minipage}}
\title{ARMA Cell: A Modular and Effective Approach for Neural Autoregressive Modeling}

\author[1]{Philipp Schiele}
\author[1]{Christoph Berninger}
\author[1]{David Rügamer}
\affil[1]{Department of Statistics, Ludwig-Maximilians-Universität München}

\begin{document}
\maketitle

\begin{abstract}
    The autoregressive moving average (ARMA) model is a classical, and arguably
    one of the most studied approaches to model time series data. It has
    compelling theoretical properties and is widely used among practitioners.
    More recent deep learning approaches popularize recurrent neural networks
    (RNNs) and, in particular, Long Short-Term Memory (LSTM) cells that have
    become one of the best performing and most common building blocks in neural
    time series modeling. While advantageous for time series data or sequences
    with long-term effects, complex RNN cells are not always a must and can
    sometimes even be inferior to simpler recurrent approaches. In this work, we
    introduce the ARMA cell, a simpler, modular, and effective approach for time
    series modeling in neural networks. This cell can be used in any neural
    network architecture where recurrent structures are present and naturally
    handles multivariate time series using vector autoregression. We also
    introduce the ConvARMA cell as a natural successor for spatially-correlated
    time series. Our experiments show that the proposed methodology is
    competitive with popular alternatives in terms of performance while being
    more robust and compelling due to its simplicity.
\end{abstract}

\newpage
\tableofcontents
\newpage

\section{Introduction}

Despite the rapidly advancing field of deep learning (DL), linear autoregressive
models remain popular for time series analysis among academics and
practitioners. Especially in economic forecasting, datasets tend to be small and
signal-to-noise ratios low, making it difficult for neural network approaches to
effectively learn linear or non-linear patterns. Although research in the past
has touched upon autoregressive models embedded in neural networks
(\eg, \cite{connor1991recurrent,connor1994}), existing literature in fields
guided by linear autoregressive models such as econometrics mainly focuses on
hybrid approaches (see Section~\ref{sec:related}). These hybrid approaches
constitute two-step procedures with suboptimal properties and often cannot even
improve over the pure linear model. The DL community took a different route for
time-dependent data structures, popularizing recurrent neural networks (RNNs),
as well as adaptions to RNNs to overcome difficulties in training and the
insufficient memory property~\cite{Hochreiter.1997} of simpler RNNs. In
particular, methods like the Long Short-Term Memory (LSTM) cell are frequently
used in practice, whereas older recurrent approaches such as Jordan or Elman
networks seem to have lost ground in the time series modeling
community~\cite{jordan:attractor, ELMAN1990179}. This can be attributed to the
more stable training and the insensitivity to information lengths in the data of
more recent recurrent network approaches such as the LSTM cell.

While often treated as a gold standard, we argue that these more complex RNN
cells (such as the LSTM cell) are sometimes used only because of the lack of
modular alternatives and that their long-term dependencies or data-driven forget
mechanisms might not always be required in some practical applications. For
example, in econometrics, including a small number of lagged time series values
or lagged error signals in the model is usually sufficient to explain most of
the variance of the time series. Similar, sequences of images (\ie,
tensor-variate time series) such as video sequences often only require the
information of a few previous image frames to infer the pixel values in the next
time step(s). In addition, current optimization routines allow practitioners to
train classical RNN approaches without any considerable downsides such as
vanishing or exploding gradients.

\begin{figure*}
    \centering
    \includegraphics[width=0.9\textwidth]{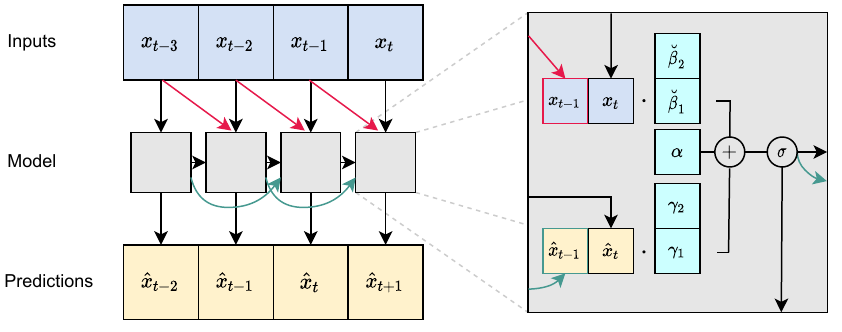}
    \caption{Left: Graphical visualizations of how predictions are computed in a
        univariate ARMA($2$,$2$) cell using the time series values $x$ from the
        current and previous time points as well as past model predictions
        $\hat{x}$. Right: Zooming in on the rightmost model cell from the left
        picture to show the computations of the ARMA cell with parameters as
        defined in~\eqref{eq:arma_reform}.}
    \label{fig:overview}
\end{figure*}

\paragraph{Our contributions.}

In this work, we propose a new type of RNN cell (cf. Figure~\ref{fig:overview})
that can be seen as a natural connection between the classical time series
school of thoughts and DL approaches. To analyze how the ARMA modeling
philosophy can improve neural network predictions, we
\begin{itemize}
    \item embed ARMA models in a neural network cell, which has various
          advantages over classical approaches (see Section~\ref{sec:proposal});
    \item further exemplify how this proposal can be extended to convolutional
          approaches to model tensor-variate time series such as image sequences
          (Section~\ref{sec:conv});
    \item demonstrate through various numerical experiments that our model is
          on par with or even outperforms both its classical time series pendant
          as well as the LSTM cell in various settings, with
          architectures ranging from shallow linear to deep non-linear time
          series models;
    \item provide a fully-tested, modular and easy-to-use
          \texttt{TensorFlow}~\cite{TF.2016} implementation with a high-level
          syntax almost identical to existing RNN cells to foster its usage and
          systematic comparisons. It is available
          at \url{https://github.com/phschiele/armacell}.
\end{itemize}

The goal of this paper is further to make practitioners aware of an alternative
to commonly used RNN cells, highlight that short-term recurrence can be
sufficient in various time series applications, and that a simpler parameterized
lag structures can even outperform data-driven forget mechanisms.

We start by discussing related literature in the following section. A short
mathematical background is given in Section~\ref{sec:background}, followed by
our proposed modeling approach in Section~\ref{sec:proposal}. We investigate
practical aspects of our method in Section~\ref{sec:experiments} and summarize
all ideas and results in Section~\ref{sec:conclusion}.

\section{Related literature} \label{sec:related}

Recent advancements in (deep) time series modeling merge statistical
autoregressive approaches with DL, specifically in building larger modeling
frameworks with multiple components, including appropriate
preprocessing~\cite{Salinas.2020, han2019review}. We emphasize literature
synthesizing classical time series analysis with fundamental building blocks of
RNN modeling.

\paragraph{Traditional autoregressive approaches.}

Building upon the foundational contributions of Box et al.~\cite{Box2015},
Autoregressive Integrated Moving Average (ARIMA) models have been popularized
due to their simple design coupled with practical effectiveness. They remain
widely used in statistics, econometrics, and many other fields. In instances
where the mean stationarity of a time series is attained without necessitating
an integration step, the Autoregressive Moving Average (ARMA) model, a special
case of ARIMA, becomes applicable. A spectrum of specialized variants has been
developed to adeptly model certain nonlinear series. Notable examples include
the bilinear models by Granger \& Andersen~\cite{granger1978invertibility,rao1981theory} and
threshold models by Tong \& Lim~\cite{tong2009threshold}. In addition, Seasonal ARIMA
(SARIMA) models can include seasonal patterns, and the ARIMAX models have been
extended to incorporate exogenous variables~\cite{Box2015}. Analogies between
ARMA models and RNNs have been discussed by various
authors~\cite{connor1991recurrent,connor1994,saxen1997equivalence}, which we
will discuss in the following.

\paragraph{Recurrent neural network approaches.}
RNNs naturally extend traditional models by incorporating previous states into
current predictions. Notable variants such as Elman~\cite{ELMAN1990179} and
Jordan~\cite{jordan:attractor} networks have been developed, but suffer from
issues like vanishing or exploding gradients (\eg, \cite{GoodBengCour16})
Advances like Gated Recurrent Units (GRU)~\cite{cho2014properties} and Long
Short-Term Memory cells (LSTM)~\cite{Hochreiter.1997} have been instrumental in
mitigating these challenges, facilitating the modeling of long-term
dependencies.

\paragraph{Combining classical time series approaches with neural networks.}
Hybrid models combining classical and neural network approaches have been
explored, often involving a two-stage process where a neural network refines the
residuals of a preceding AR(I)MA model \cite{zhang2003time}. While several
combinations, such as state space models with RNNs, have been proposed, they
tend to lack modularity and general applicability \cite{rivals1996black}.

\paragraph{Recurrent convolutional approaches.}

Spatio-temporal sequence forecasting benefits from the incorporation of
convolutional operations into RNN cells, as seen in models like ConvLSTM and
convolutional GRU which handle spatially distributed information efficiently
\cite{NIPS2015_07563a3f, tian2019generative}.

\section{Background and notation} \label{sec:background}

In the following, we introduce our notation and the general setup for modeling
time series. We will address univariate time series $x_t \in \reals$ for
time points $t \in \integers$ as well as
multi- and tensor-variate time series, which we denote as $\boldsymbol{x}_t$ and
$\boldsymbol{X}_t$, respectively.

\paragraph{ARMA model.}

The ARMA($p, q$) model~\cite{Box2015} for $p,q \in \naturals_0$ is defined as
\begin{equation}
    \label{eq:arma}
    x_t = \alpha + \sum_{i=1}^p  \beta_i x_{t-i} + \sum_{j=1}^q  \gamma_j\varepsilon_{t-j} + \varepsilon_t,
\end{equation}
where $x_t$ represents the variable of interest defined for $t \in \integers$
and is observed at time points $t = 1, \dots,T$, $T \in \naturals$\footnote{It
    is common to define a time series for time points $t=1,\ldots,T$ to describe its
    current value and recent history, while time series dynamics are assumed to
    originate from time points prior to $t=1$, hence $t\in\integers$}. Here, $\alpha$,
$\beta_1,\ldots,\beta_p$, $\gamma_1,\ldots,\gamma_q$ are real valued parameters
and $\varepsilon_t \overset{iid}{\sim} \mathcal{F}(\sigma^2)$ is an independent
and identically distributed (iid) stochastic process with pre-specified
distribution $\mathcal{F}$ and variance parameter $\sigma^2>0$.
By setting $q=0$ or $p=0$, the ARMA model comprises the special cases of a pure
autoregressive (AR) and a pure moving average (MA) model, respectively. The
class of ARMA models is, in turn, a special case of ARIMA models, where
differencing steps are applied to obtain a stationary mean function before
fitting the ARMA model. As stationarity is also a fundamental assumption for
RNNs to justify parameter sharing~\cite{GoodBengCour16}, we focus on the class
of ARMA models in this work, \ie, assume that differencing has already been
applied to the data.  A stationary time
series is characterized by a constant mean and variance and a time invariant
autocorrelation structure. $\alpha_0, \dots , \alpha_p$ and $\beta_1, \dots,
    \beta_q$ are model parameters and $p$ and $q$ characterize the number of lags of
the dependent variable and the forecasting errors included in the model,
respectively.

\paragraph{VARMA model.}
The univariate ARMA model can be generalized to a multivariate version -- the
vector autoregressive moving average (VARMA) model -- by adapting the principles
of the ARMA model for multivariate time series. The VARMA($p,q$)
model~\cite{tiao1981} for $p,q \in \naturals_0$ is defined as
\begin{equation}
    \label{eq:varma}
    \boldsymbol{x}_t = \boldsymbol{\alpha} + \sum_{i=1}^p \boldsymbol{B}_i \boldsymbol{x}_{t-i} + \sum_{j=1}^q
    \boldsymbol{\Gamma}_j \boldsymbol{\varepsilon}_{t-j} + \boldsymbol{\varepsilon}_t
\end{equation}
where $\boldsymbol{x}_t, t\in\integers$ represents a vector of time series
observed at time points $t = 1, \dots,T$. $\boldsymbol{B}_i$  and
$\boldsymbol{\Gamma}_j$ are time-invariant $(k \times k)$-matrices, where
$k\in\naturals$ represents the number of individual time series.
$\boldsymbol{\varepsilon}_t$ is a $k$-dimensional iid stochastic process with
pre-specified $k$-dimensional distribution $\mathcal{F}(\boldsymbol{\Omega})$
and covariance matrix $\boldsymbol{\Omega}$.
Furthermore, the error term $\boldsymbol{\varepsilon}_t$ needs to satisfy the
following three conditions: \begin{enumerate} \item
$E[\boldsymbol{\varepsilon}_t] = \boldsymbol{0}$ for all $t\in\mathcal{T}$
\item $E[\boldsymbol{\varepsilon}_t\boldsymbol{\varepsilon}_t^T] =
\boldsymbol{\Omega}$ is a (k x k) positive semi-definite covariance matrix
\item $E[\boldsymbol{\varepsilon}_t\boldsymbol{\varepsilon}_{t-c}^T] =
\boldsymbol{0}$ for any non-constant $c \in\naturals$. \end{enumerate}
By setting $q=0$, the VARMA model comprises the special cases of a pure
autoregressive (VAR) model, which is the most common VARMA model used in
applications. Similar to the ARMA model being a special case of the ARIMA model
class, the VARMA model is a special case of the VARIMA model class, representing
only stationary time series.

\section{ARMA-based neural network layers} \label{sec:proposal}

ARMA models have been successfully used in many different fields and are a
reasonable modeling choice for time series in many areas. This section
introduces a neural network cell version of the ARMA mechanism. While very
similar to Elman or Jordan networks, the proposed cell exactly resembles the
ARMA computations and can be used in a modular fashion in any neural network
architecture where recurrent structures are present. Emulating the ARMA logic in
a recurrent network cell has various advantages. It allows to
1) recover estimated coefficients of classical ARMA software (see Supplementary
Material~\ref{app:paramrecov} for an empirical investigation of the
convergence), but can also be used to fit ARMA models for large-scale or
tensor-variate data (which is otherwise computationally infeasible), 2)
modularly use the ARMA cell in place for any other RNN cell, 3) combine ARMA
approaches with other features from neural networks such as regularization and
thereby seamlessly extend existing time series models, and 4)
model hybrid linear and deep network models that were previously only possible
through multi-step procedures. As shown in our numerical experiments section, an
ARMA cell can further lead to comparable or even better prediction performance
compared to modern RNN cells.

\subsection{ARMA cell} \label{sec:cell}

An alternative formulation of the ARMA model can be derived by incorporating the
observed (or estimated) residual term $\hat{\varepsilon}$ through the
predictions $\hat{x}_t \coloneqq x_t - \hat{\varepsilon}_t, t\in\integers$.
Thus,~\eqref{eq:arma} can be defined in terms of its intercept, the model
predictions $\hat{x}_t$ and the actual time series values $x_t$ as
\begin{equation}
    \label{eq:arma_reform}
    \hat{x}_t = \alpha + \sum_{i=1}^{\max(p,q)} \breve{\beta}_i x_{t-i}  - \sum_{j=1}^q \gamma_j \hat{x}_{t-j},
\end{equation}
where
\[
    \breve{\beta}_i = \begin{cases}
        \beta_i + \gamma_i & \mbox{ for } i \leq \min(p,q),         \\
        \beta_i            & \mbox{ for } i > q \text{ and } p > q, \\
        \gamma_i           & \mbox{ for } i > p \text{ and } p < q.
    \end{cases}
\]

A detailed derivation of~\eqref{eq:arma_reform} is given in~\ref{app:derivation}.
Using~\eqref{eq:arma_reform}, we can implement the ARMA functionality as an RNN
cell. More specifically, the recurrent cell processes the $p$ lagged time series
values as well as the $q$ predicted outputs of the previous time steps and
computes a linear combination with parameters $\breve{\beta}_{i}$ and
$\gamma_{j}$. After adding a bias term, the final output $\hat{x}_t$ is given by
a (non-linear) activation function $\sigma$ of the sum of all terms.
Figure~\ref{fig:overview} gives both a higher-level view of how predictions are
computed in the ARMA cell as well as a description of how the cell is defined in
detail. In addition to the classical ARMA computations in the cell, the
activation function $\sigma$ allows to straightforwardly switch between a linear
ARMA model and a non-linear version.

The above-mentioned ARMA cell has the same hypothesis space as the classical
ARMA model when using a single-unit ARMA cell with a linear activation function.
While using a non-linear activation for the outputs, in this case, is equivalent
to using a link function (as done in generalized linear models) for the
classical ARMA model, extensions using multiple units or stacking ARMA cells
(see below) increase the model's expressiveness. As for regular multi-layer
perceptrons, where each node is a simple regression model with an activation
function and the combination of multiple units makes the models more expressive,
these extensions combine simpler ARMA models and therefore allow modeling more
complex relationships.

\paragraph{Advantages and comparison to other cells.}

Modeling a classical ARMA model in a neural network can be more stable in the
estimation of coefficients due to the use of a stochastic first-order method
(less vulnerable to ill-conditioning and numerical instabilities), which is also
confirmed in our numerical experiments in numerous settings. Training the ARMA
model using mini-batch optimization, further allows scaling to large data sets,
which is especially beneficial when modeling multivariate time series where the
complexity of classical models increases substantially with the number of
parameters and the multivariate time series dimension.

In contrast to the standard RNN cell, the ARMA cell internally can access
multiple previous states and lagged features, making it potentially easier to
learn time dependencies and recurrences. The standard RNN cell, in contrast,
only relies on the current input and the previous cell state. In other words,
the ARMA cell allows for a more complex autoregressive structure and, in
contrast to the simple RNN, provides a way to model moving averages. This can
also be explained using Figure~\ref{fig:overview}, where the standard RNN cell can represent the
black arrows, but not the red and green connections.

Last but not least, the ARMA cell can be used to seamlessly model hybrid models
end-to-end in one holistic network, which historically has always been
implemented using two-step approaches \cite{zhang2003time}, yielding potentially
inferior performance as the models in both steps are not jointly optimized. %
This is also confirmed by our numerical results in Section~\ref{sec:hybrid}.

\subsection{Training procedure}

The ARMA cell is trained as follows. For a given sequence, it creates
predictions by recursively applying \eqref{eq:arma_reform}. This is done in one
forward pass. To also allow predictions for the first $q$ time points in each
sequence, we need to pad the sequence of previous predictions with 0-values.
Further details on the input sizes of the ARMA cell can be found in
Supplementary Material. We then differentiate the loss of these outputs given
the current weights back through the whole sequence, \ie, the network is
trained exactly as done for the LSTM, GRU, and simple RNN cell via
backpropagation through time. Note that our ARMA cell also supports returning
sequences, which we can use to stack cells or for training a model on multiple
steps simultaneously.

\subsection{Extensions}

The ARMA cell in Figure~\ref{fig:overview} can be used in a modular fashion
similar to an LSTM or GRU cell. In the following, we will thus present how this
idea can be used to generate more complex architectures using multiple units or
by stacking cells. Both options also allow bypassing the linearity assumptions
of ARMA models.

\begin{figure} \centering
    \includegraphics[width=0.5\textwidth]{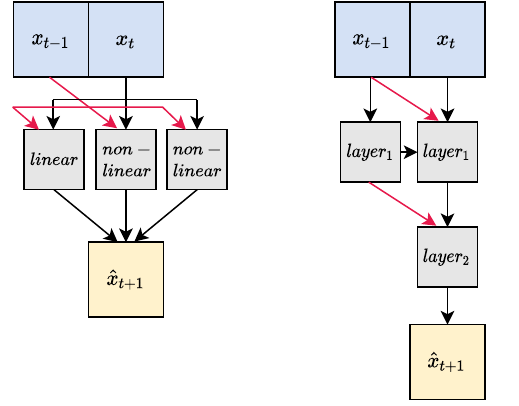}
    \caption{Visualization of an ARMA cell with multiple units representing
    a mixture of linear and non-linear ARMA models by using different
    activation functions (left) and a network with stacked ARMA cells
    creating a more complex model class by transforming inputs by subsequent
    ARMA cells (right).} \label{fig:multi} \end{figure}

\paragraph{Multi-unit ARMA cell.}
Similar to feedforward neural networks, an RNN layer can also contain multiple
units. Each unit receives the same input but can capture different effects due
to the random initialization of weights. The outputs of each unit are then
concatenated. A multi-unit architecture allows combining different activation
functions, \eg, to simultaneously capture linear and non-linear effects, and is
depicted in Figure~\ref{fig:multi} (left). Using a multi-unit ARMA cell thereby
seamlessly provides the possibility to combine a linear with a non-linear ARMA
model. We refer to models having a single hidden ARMA layer with one or more
units as \textit{ShallowARMA} models.

\paragraph{Stacked ARMA.}
To allow for higher levels of abstraction and increased model complexity, the
ARMA modeling strategy does not only allow for multiple units in a single layer,
but users can also stack multiple layers in series, as shown in
Figure~\ref{fig:multi} (right). This is achieved by returning a sequence of
lagged outputs from the previous layer. Models with more than one hidden ARMA
layer are referred to as \textit{DeepARMA} models in the following.

\subsection{ConvARMA} \label{sec:conv}

\begin{figure} \centering \includegraphics[width=0.7\textwidth]{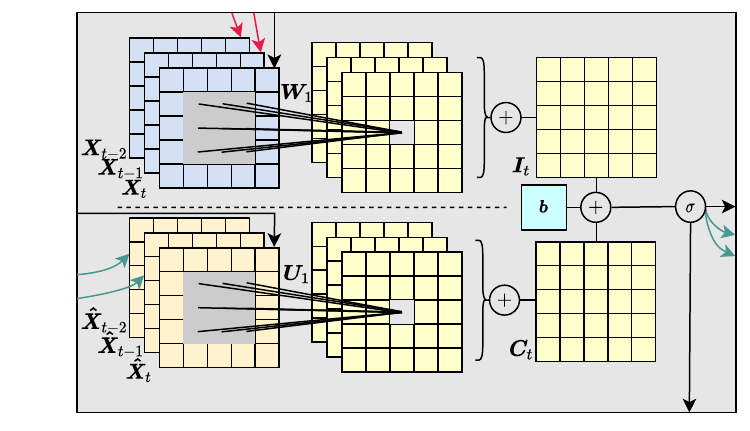}
      \caption{Exemplary visualization of a single-filter ConvARMA cell
      processing matrix-variate time series (with a single channel) with three
      lags (upper left) and matrix-variate predictions with three lags (bottom
      left) using convolutions and combining the results into a single matrix
      prediction (bottom/top right) with additional bias term $b$ and
      activation function $\sigma$ (center right).} \label{fig:convarma}
      \end{figure}

Similar to the ConvLSTM network~\cite{NIPS2015_07563a3f}, it is possible to
model spatial dependencies and process tensor-variate time series $\boldsymbol{X}_t \in
    \reals^{n_1 \times \ldots \times n_d}, n_1,\ldots,n_d \in \naturals, d \in
    \naturals$ by using convolution operations within an ARMA cell. The resulting
ConvARMA($p,q$) cell for $p,q \in \naturals_0$ and $t\in\integers$ is defined
as
\begin{align}
    \label{eq:convarma}
    \boldsymbol{I}_t       & = \sum_{i=1}^p \boldsymbol{W}_i * \boldsymbol{X}_{t-i}, \notag                       \\
    \boldsymbol{C}_t       & = \sum_{j=1}^q \boldsymbol{U}_j * \hat{\boldsymbol{X}}_{t-j},                        \\
    \hat{\boldsymbol{X}}_t & = \sigma \left( \boldsymbol{I}_t + \boldsymbol{C}_t + \boldsymbol{b} \right), \notag
\end{align}
where $*$ represents the convolution operator, $\boldsymbol{W}_i \in
    \reals^{k_1 \times \ldots \times k_{d-1} \times n_d \times c}, i=1,\ldots,p$
and $\boldsymbol{U}_j \in \reals^{k_1 \times \ldots \times k_{d-1}\times c
    \times c}, j=1,\ldots,q$ are the model's kernels of size $k_1\times\ldots\times
    k_{d-1}$, $\boldsymbol{b}\in\reals^{c}$ is a bias term broadcasted to
dimension $n_1 \times \ldots \times n_{d-1}\times c$ and $\sigma$ an activation
function. By convention, the last dimension of the input represents the
channels, and $c$ denotes the number of filters of the convolution. The inputs
of the convolution are padded to ensure that the spatial dimensions of the
prediction $\hat{\boldsymbol{X}}_t$ and the state remain unchanged. In other
words, the ConvARMA cell resembles the computations of an ARMA model, but
instead of simple multiplication of the time series values with scalar-valued
parameters, a convolution operation is applied. Figure~\ref{fig:convarma} shows
an abstract visualization of the computations in a ConvARMA cell.
To follow the AR(I)MA modeling logic in the spatial dimensions, a ConvARMA cell
can further incorporate spatial differences in all directions.
A possible extension of the cell proposed in~\eqref{eq:convarma} could further
be to allow for non-linear recurrent activations as done for, \eg, the ConvLSTM
cell.

As for the ConvLSTM or ConvGRU cell, the ConvARMA cell can be included in an
autoencoder architecture for sequence-to-sequence modeling or extended to \eg,
allow for warping, rotation, and scaling~\cite{shi2017deep}.

\subsection{Limitations} As for other autoregressive approaches, our approach is
limited in its application if the time series are very short or if a large
number of lags $p$ is required to approximate the underlying data generating
process well. We note, however, that due to the model's recurrent definition,
past time points $t-i$ for $i>p$ also influence the model's predictions. It is
therefore often not necessary to define a large lag value $p$, even if
autocorrelation is high. Despite the ARMA cell's simplicity, this also shows
that its predictions are not always straightforward to interpret.

\section{Numerical experiments} \label{sec:experiments}

In this section, we examine the performance of our ARMA cell in a variety of
synthetic and benchmark experiments. We examine how it compares to classical
time series approaches as well as to similar complex neural architectures. Note
that our experiments are not designed to be a benchmark comparison with current
state-of-the-art time series forecasting frameworks. These rather complex
architectures include many different components, such as automated
pre-processing and feature generation, and thus do not necessarily allow making
a statement about the performance of a single recurrent cell therein. Instead,
we aim for a comparison with other fundamental modeling building blocks
\footnote{We provide code to reproduce all experiments at
\url{https://github.com/phschiele/armacell_paper}}. Yet, in order to emphasize
our cell's modularity and demonstrate its efficacy when used as part of a larger
state-of-the-art network, we also present results on real-world benchmark data
sets when replacing RNN cells within a DeepAR model \cite{Salinas.2020} with the
ARMA cell.

\paragraph{Methods.} For (multivariate) time series, we compare a shallow and a
deep variant of the ARMA cell against the respective (V)ARMA model and neural
models. For the latter, we consider LSTM, GRU, and simple RNN cells, again each
in their shallow and deep variants. Hyperparameter optimization is done using
a grid search with predefined parameter spaces for the number of units for all
network layers and lags for ARMA-type models. All other hyperparameters of
network layers are kept fixed with defaults that do not favor one or the other
method.
Further details on the specification of the architectures can be found in the
Supplementary Material.

\paragraph{Performance measures.} We compare time series predictions using the
root mean squared error (RMSE) for both the uni- and multivariate time series
forecasts. We provide further performance measures for our comparisons in the
Supplementary Material.

\paragraph{Comparison between classical and first-order optimization.} In the
case where the data generating process is in fact a (V)ARMA process, we expect
the classical (V)ARMA model and the ARMA cell to perform similarly, but note
that the optimization using stochastic gradient descent can sometimes yield
better estimations of this process and hence outperform these classical models
despite having the exact same hypothesis space.

\subsection{Simulation study.}

We start with a variety of synthetic data examples using time series models
defined in Lee et al.~\cite{lee1993testing}. Simulations include linear and non-linear, as
well as uni- and multivariate time series. All time series are of length 1000
and split into 70\% train and 30\% test data. The data generating processes
follow Lee et al.~\cite{lee1993testing} and include an ARMA process (ARMA), a threshold
autoregressive model (TAR), an autoregressive time series which is transformed
using the sign operation (SGN), a non-linear autoregressive series (NAR), a
heteroscedastic MA process (Heteroscedastic), a vector ARMA (VARMA), a
non-linear multivariate time series with quadratic lag structure (SQ), and an
exponential multivariate autoregressive time series (EXP). The exact
specification of the data generating processes can be found in the Supplementary
Material.

\begin{table*}[t]
    \caption{Comparisons of different methods (rows) and different data generating processes (columns) using the average RMSE $\pm$ the standard deviation of 30 independent runs. The best performing method is highlighted in bold, the second-best in italics.}
    \label{simulation_merged}
    \begin{center}
        \begin{small}
            \resizebox{\textwidth}{!}{
                \begin{tabular}{llllll||lll}
                    \toprule
                    {}                      & \multicolumn{5}{c||}{Univariate} &
                    \multicolumn{3}{c}{Multivariate}                                                                                                          \\
                    {}                      & ARMA                             & TAR                     & SGN           & NAR & Heterosk. & VARMA & EXP & SQ
                    \\
                    \midrule
                    $\text{(V)ARMA}$        & 2.04$\pm$0.35                    &
                    2.92$\pm$3.20           & 2.36$\pm$2.67                    & 2.67$\pm$4.92
                                            & 1.19$\pm$0.15                    & \bfseries 1.00$\pm$0.03 & 3.35$\pm$0.69
                                            & 1.90$\pm$0.14                                                                                                   \\
                    $\text{ShallowARMA}$    & \bfseries 1.96$\pm$0.09          & \bfseries
                    1.09$\pm$0.12           & 1.10$\pm$0.07                    & \itshape 1.02$\pm$0.05  &
                    \bfseries 1.11$\pm$0.06 & \itshape 1.00$\pm$0.03           & \itshape
                    3.14$\pm$0.74           & \bfseries 1.75$\pm$0.12                                                                                         \\
                    $\text{DeepARMA}$       & 1.97$\pm$0.10                    & \itshape
                    1.24$\pm$0.47           & \bfseries 1.05$\pm$0.06          & 1.02$\pm$0.04           &
                    \itshape 1.11$\pm$0.06  & 1.01$\pm$0.03                    & \bfseries
                    3.10$\pm$0.75           & \itshape 1.76$\pm$0.11                                                                                          \\
                    $\text{LSTM}$           & 1.98$\pm$0.10                    &
                    1.38$\pm$0.42           & 1.19$\pm$0.16                    & 1.02$\pm$0.04
                                            & 1.15$\pm$0.09                    & 1.01$\pm$0.04           &
                    3.26$\pm$0.73           & 1.83$\pm$0.16                                                                                                   \\
                    $\text{DeepLSTM}$       & 2.02$\pm$0.10                    &
                    1.46$\pm$0.55           & 1.17$\pm$0.13                    & 1.02$\pm$0.04
                                            & 1.16$\pm$0.08                    & 1.03$\pm$0.04           &
                    3.35$\pm$0.68           & 1.86$\pm$0.14                                                                                                   \\
                    $\text{GRU}$            & \itshape 1.96$\pm$0.10           &
                    1.28$\pm$0.31           & \itshape 1.09$\pm$0.08           & \bfseries
                    1.02$\pm$0.04           & 1.13$\pm$0.07                    & 1.02$\pm$0.04           &
                    3.19$\pm$0.75           & 1.80$\pm$0.13                                                                                                   \\
                    $\text{DeepGRU}$        & 1.99$\pm$0.09                    &
                    1.24$\pm$0.36           & 1.09$\pm$0.12                    & 1.02$\pm$0.04
                                            & 1.12$\pm$0.06                    & 1.01$\pm$0.04           &
                    3.22$\pm$0.80           & 1.82$\pm$0.18                                                                                                   \\
                    $\text{Simple}$         & 1.99$\pm$0.09                    &
                    1.29$\pm$0.31           & 1.14$\pm$0.09                    & 1.04$\pm$0.04
                                            & 1.13$\pm$0.08                    & 1.02$\pm$0.03           &
                    3.29$\pm$0.70           & 1.80$\pm$0.12                                                                                                   \\
                    $\text{DeepSimple}$     & 2.01$\pm$0.11                    &
                    1.47$\pm$0.57           & 1.15$\pm$0.10                    & 1.03$\pm$0.04
                                            & 1.16$\pm$0.10                    & 1.02$\pm$0.03           &
                    3.30$\pm$0.84           & 1.83$\pm$0.13                                                                                                   \\
                    \bottomrule
                \end{tabular}
            }
        \end{small}
    \end{center}
    \vskip -0.1in
\end{table*}

\paragraph{Results.} The results in Table~\ref{simulation_merged} suggest that
the ShallowARMA approach emulating an ARMA model in a neural network works well
for all linear- and non-linear datasets. In terms of robustness, the lower RMSE
and high standard deviation of the ARMA model on the ARMA process shows that
fitting an ARMA model in a neural network with stochastic gradient descent can,
in fact, be more robust than the standard software~\cite{autoarima,
    seabold2010statsmodels}. While the classical ARMA did match the performance of
its neural counterpart it some cases, the average RMSE is worse, as it did not
converge in all runs, even for the linear time series. The performance of the
DeepARMA approach is slightly worse compared to the ShallowARMA in most cases.
The performance of LSTM, GRU, and the Simple RNN are all similar, with all
methods matching the ARMA cells in some cases, and falling slightly behind in
others. As expected, the classical ARMA approach does not work well for
non-linear data generating processes (TAR, SGN, NAR) and yields unstable
predictions underpinned by the large standard deviations in RMSE values.

For multivariate time series results of the simulation are summarized in
Table~\ref{simulation_2}. The results again show that the ShallowARMA model
matches the performance of the classical VARMA model for a dataset that is also
based on a VARMA process. For other types of data generation, the ShallowARMA
model and DeepARMA model work similarly well. Both outperform the other neural
cells, which in turn yield better results than the VARMA baseline.

In summary, the findings suggest that ARMA cells work well both for simpler
linear as well as non-linear data generating processes while being much more
stable than a classical ARMA approach. In the Supplementary Material, we further
study the empirical convergence of a single unit single hidden layer ARMA cell,
which is mathematically equivalent to an ARMA model for given values of $p$ and
$q$, and present another comparison showing the equivalence of the ARMA cell and
an Elman network.

\subsection{Ablation studies}

To explore the validity of our presented results, we perform a series of
ablation studies. Two important influence factors on the performance of time
series models are the length of the time series and the forecasting horizon,
which we subsequently assess in the controlled setting of our simulation study.

\subsubsection{Time series length}
In order to investigate the influence of the time series length on the
performance reported in previous simulations, we vary $T\in\{200, 1000, 10000\}$
and re-run the experiments reported in Table~\ref{simulation_merged}. The full
results are given in the Supplementary Material. In summary, the rank of the
different methods is similar to the aforementioned results, and ShallowARMA
yields the best results in most settings. There is, however, a clear trend in
that the performance differences between the different cell types become
irrelevant for an increase in $T$. For example, for the multivariate time series
study setting SQ, the ShallowARMA yields a notably better MSE for $T=200$
compared to the DeepSimple cell ($1.97\pm 0.41$ vs.~$3.00\pm 3.95$), the
performances are almost identical for $T=10,000$ observations ($1.78\pm 0.05$
vs.~$1.80 \pm 0.07$).

\subsubsection{Forecasting horizon}
Similar to the previous ablation study, we reran the experiments but now alter
the forecasting horizon by comparing a one-step, 10-step, and 20-step forecast
for $T=1000$. The full results can again be found in the Supplementary Material.
In the univariate case for forecasting horizons greater one, the different ARMA
variations do not outperform other approaches anymore and DeepLSTM, GRU, or
DeepGRU yield the best results in many cases. The performance values, however,
are in most cases within one standard deviation of those by the Shallow- or
DeepARMA approach. For the multivariate case, the classical VARMA model provides
the best forecast for all multi-step ahead forecast scenarios, closely followed
by the Shallow- and DeepARMA models.

\subsection{Comparison to hybrid models} \label{sec:hybrid}

We now investigate the differences between a standard hybrid approach
following~\cite{zhang2003time} and an end-to-end approach using the ARMA cell.
The hybrid model first trains a classical model, in this case, an
ARMA($2$,$2$) model. Then, an LSTM model is fit on the residual. The final
prediction is obtained as the sum of the ARMA and LSTM predictions. We also
implement an end-to-end version of this model using the ARMA cell, which we
refer to as End2End. Here, we train a linear ARMA cell, also specified with
$p=2$ and $q=2$, and sum its output with the output from an LSTM cell. Both
model approaches have the same hypothesis space, however, training a single
model simplifies the training process and optimized the parameters jointly. We
run both approaches on the simulated univariate  time series, as shown in
Table~\ref{tab:hybrid}. We see that the End2End model performs similarly to
the hybrid model in most cases. However, for the NAR dataset, we find that the
two-step hybrid approach does not converge in all cases, leading to a
substantially worse average RMSE and a high corresponding standard deviation,
indicating that this approach is less robust compared to the End2End model.

\begin{table}[] \centering

\resizebox{1.0\textwidth}{!}{ \begin{tabular}{lccccc} \toprule {} &  ARMA &
TAR &  SGN &  NAR &  Heteroskedastic \\
\midrule End2End &     2.021 $\pm$ 0.105 &  \bfseries  1.134 $\pm$ 0.127 &
\bfseries 1.128 $\pm$ 0.055 & \bfseries 1.002 $\pm$ 0.044 &  \bfseries 1.142
$\pm$ 0.062 \\
Hybrid  &  \bfseries  2.014 $\pm$ 0.112 &    1.264 $\pm$ 0.811&   1.138 $\pm$
0.051 &    3.009 $\pm$ 8.869 &  1.147 $\pm$ 0.062 \\
\bottomrule \end{tabular}
}

    \caption{Comparisons of the End2End and Hybrid approach on different data generating processes (columns) for
    univariate time series using the average RMSE $\pm$ the standard deviation of 30 independent runs. The best-performing method is highlighted in bold.}
    \label{tab:hybrid}
\end{table}

\subsection{Benchmarks}

In order to investigate the performance of our approach for real-world time
series with a potentially more complex generating process, we compare the
previously defined models on various time series benchmark datasets.

\subsubsection{Univariate and multivariate time series}

We use the m4~\cite{Makridakis.2018}, traffic~\cite{yu.2016}, and
electricity~\cite{yu.2016} dataset, all openly accessible and commonly used in
time series forecast benchmarks. Further background on every dataset and details
on pre-processing can be found in the Supplementary Material. As all datasets
come with multiple time series, we use these datasets both for testing the
performance on univariate and multivariate time series. For univariate time
series, this is done by training a local model for every dimension and averaging
the results over the different multivariate dimensions. The multivariate
comparison is based on the predictions of a single global model.

\begin{table}[t]
    \caption{Comparison of different univariate and multivariate forecasting
        approaches (rows) for different datasets (columns) based on the average
        RMSE $\pm$ the standard deviation of 10 independent runs. The best
        performing method is highlighted in bold, the second-best in italics.}
    \label{t-benchmark}
    \begin{center}
        \begin{small}
            \begin{sc}

                    \begin{tabular}{clccc}
                        {}            &                         & m4                      & traffic
                                      & elec.                                                                        \\
                        \midrule
                        \parbox[t]{2mm}{\multirow{9}{*}{\rotatebox[origin=c]{90}{univ.\,\,}}}
                                      & $\text{ARMA}$           & 1.58$\pm$0.00           &
                        0.98$\pm$0.00 & 1.19$\pm$0.00                                                                \\
                                      & $\text{ShallowARMA}$    & \bfseries 1.57$\pm$0.01 &
                        0.97$\pm$0.00 & 1.14$\pm$0.01                                                                \\
                                      & $\text{DeepARMA}$       & \itshape 1.57$\pm$0.01  & \bfseries
                        0.94$\pm$0.01 & \bfseries 1.10$\pm$0.02

                        \\
                                      & $\text{LSTM}$           & 1.71$\pm$0.14           & \itshape
                        0.96$\pm$0.01 & 1.15$\pm$0.07                                                                \\
                                      & $\text{DeepLSTM}$       & 1.96$\pm$0.38           &
                        0.97$\pm$0.02 & 1.12$\pm$0.05                                                                \\
                                      & $\text{GRU}$            & 1.61$\pm$0.03           &
                        0.97$\pm$0.02 & \itshape 1.11$\pm$0.02                                                       \\
                                      & $\text{DeepGRU}$        & 1.61$\pm$0.02           &
                        0.97$\pm$0.02 & 1.11$\pm$0.03                                                                \\
                                      & $\text{Simple}$         & 1.72$\pm$0.15           &
                        1.00$\pm$0.01 & 1.12$\pm$0.01                                                                \\
                                      & $\text{DeepSimple}$     & 1.76$\pm$0.17           &
                        1.00$\pm$0.02 & 1.11$\pm$0.02                                                                \\
                        \toprule
                        \midrule
                        \parbox[t]{2mm}{\multirow{9}{*}{\rotatebox[origin=c]{90}{multiv.\,\,}}}
                                      & $\text{ARMA}$           & 1.72$\pm$0.00           & \itshape 1.06$\pm$0.00 &
                        1.46$\pm$0.00                                                                                \\
                                      & $\text{ShallowARMA}$    & \itshape 1.68$\pm$0.01  & \bfseries
                        1.06$\pm$0.00 & 1.37$\pm$0.03                                                                \\
                                      & $\text{DeepARMA}$       & \bfseries 1.67$\pm$0.01 &
                        1.08$\pm$0.01 & 1.32$\pm$0.03                                                                \\
                                      & $\text{LSTM}$           & 1.92$\pm$0.15           &
                        1.15$\pm$0.01 & 2.07$\pm$1.12                                                                \\
                                      & $\text{DeepLSTM}$       & 2.11$\pm$0.25           &
                        1.15$\pm$0.02 & 1.26$\pm$0.05                                                                \\
                                      & $\text{GRU}$            & 1.91$\pm$0.25           &
                        1.15$\pm$0.00 & 1.25$\pm$0.04                                                                \\
                                      & $\text{DeepGRU}$        & 1.88$\pm$0.12           &
                        1.15$\pm$0.01 & \itshape 1.23$\pm$0.02                                                       \\
                                      & $\text{Simple}$         & 1.89$\pm$0.06           &
                        1.16$\pm$0.00 & 1.25$\pm$0.03                                                                \\
                                      & $\text{DeepSimple}$     & 1.90$\pm$0.08           &
                        1.16$\pm$0.01 & \bfseries 1.22$\pm$0.01                                                      \\
                        \bottomrule
                    \end{tabular}
            \end{sc}
        \end{small}
    \end{center}
    \vskip -0.1in
\end{table}

\paragraph{Univariate time series.} Results of univariate benchmarks are
summarized in Table~\ref{t-benchmark}. The comparisons suggest that ARMA cells,
either in the shallow or deep variant, outperform on all studied datasets. The
classical ARMA model is competitive for the m4 dataset, but again worse than its
neural pendant on Traffic, and Electricity.

\paragraph{Multivariate time series.} For the multivariate time series
benchmarks, we observe that model performance is in general worse than when
performing hyperparameter optimization and model training for each time series
individually, as done for the univariate time series benchmark. Finding
architectures better suited to the individual time series seems to outweigh the
additional information from observing the comovement of multiple time series
simultaneously. In the comparison of different forecasting approaches for
multivariate dimensions, the performance of the ARMA cells is either notably
better than the other neural cells but on par with the classical ARMA model
(Traffic), or outperforms all other approaches (m4). Only for the Electricity
dataset, the ARMA cells yield a slightly worse MSE compared to the DeepSimple
cell.

\subsubsection{Integration with state-of-the-art forecasting frameworks}
\label{sec:deepar} To demonstrate the modularity aspect of the ARMA cell, we use
it to replace the LSTM cell in a DeepAR model \cite{Salinas.2020}. We then train
a larger global model on the previously studied benchmark data sets for
multivariate time series (as this is the application area where DeepAR model
excels in performance) and examine to what extent the change in RNN cell
influences the results. The experimental details can be found in Supplementary
Material~\ref{app:deepar}.
\begin{table*}[h]
    \caption{Comparison of different DeepAR-based models (rows) for different
        datasets (columns) based on the average negative log-likelihood $\pm$
        its standard deviation of 30 independent runs. The best performing
        method is highlighted in bold.}
    \label{deepar}
    \begin{center}
        \begin{small}
            \begin{sc}
                \begin{tabular}{@{}lcccc@{}} {}            & m4
                                           & Traffic                     & Elec.     \\
               \midrule DeepAR ARMA single & 2.444 $\pm$
               0.137                       & \bfseries 1.323 $\pm$ 0.033 & \bfseries
               5.236 $\pm$ 0.398                                                     \\
               DeepAR LSTM single          & \bfseries 2.306 $\pm$ 0.056 &
               1.362 $\pm$ 0.054           & 10.236 $\pm$ 9.242                      \\ \midrule
               DeepAR ARMA stacked         & 2.513 $\pm$ 0.157           &
               \bfseries 1.332 $\pm$ 0.056 & \bfseries 5.639 $\pm$ 0.461
               \\
               DeepAR LSTM stacked         & \bfseries 2.266 $\pm$ 0.049 &
               1.381 $\pm$ 0.053           & 9.607 $\pm$  5.630                      \\
               \bottomrule
                \end{tabular}
            \end{sc}
        \end{small}
    \end{center}
    \vskip -0.1in
\end{table*}
Results (Table~\ref{deepar}) indicate that it is possible to successfully
replace the LSTM with an ARMA cell in the DeepAR model and to receive a similar
performance. We further observe that the LSTM-based DeepAR does not always
converge for the electricity dataset, indicating that the training of the ARMA
cell is more robust.

\section{Conclusion and Outlook} \label{sec:conclusion}

We provided a modular and flexible neural network cell to model time series in a
simply parameterized fashion and as an alternative to commonly used RNN cells
such as the LSTM cell. We further extended this approach to vector
autoregression and autoregressive models for tensor-variate applications. Our
numerical experiments show that the ARMA cell
1) performs well on univariate, multivariate, and tensor-variate time series;
2) matches or even outperforms the LSTM, GRU, and a simple RNN cell in linear
and non-linear settings, and; 3)
shows more robust convergence for classical ARMA formulations compared to a
standalone implementation.

\paragraph{Outlook.} An interesting future research direction is to make use of
the theoretical results for ARMA models known from classical statistical
literature and transfer these to the application of ARMA as a cell with multiple
units or in its stacked variant. A directly available result, \eg, would be
last-layer uncertainty quantification (see, \eg, \cite{immer2021improving}) in
a stacked RNN model where the last cell is an ARMA cell with one unit. Although
this neglects the variance in previous layers, it allows a first assessment of
the RNN's uncertainty. Further, when merging multiple linearly activated ARMA
cells, the combination is an ensemble of ARMA models, for which some form of
uncertainty quantification method could be derived.

\clearpage
{\small
\bibliography{refs}

\newcommand{\etalchar}[1]{$^{#1}$}
\begin{thebibliography}{SCW{\etalchar{+}}15}

\bibitem[ABC{\etalchar{+}}16]{TF.2016}
M.~Abadi, P.~Barham, J.~Chen, Z.~Chen, A.~Davis, J.~Dean, M.~Devin, S.~Ghemawat, G.~Irving, M.~Isard, et~al.
\newblock Tensor{F}low: A system for large-scale machine learning.
\newblock {\em Proceedings of the 12th USENIX Symposium on Operating Systems Design and Implementation (OSDI '16)}, pages 265--283, 2016.

\bibitem[BJRL15]{Box2015}
G.~Box, G.~Jenkins, G.~Reinsel, and G.~Ljung.
\newblock {\em Time Series Analysis: Forecasting and Control}.
\newblock John Wiley \& Sons, 2015.

\bibitem[CAM91]{connor1991recurrent}
J.~Connor, L.~Atlas, and D.~Martin.
\newblock Recurrent networks and {NARMA} modeling.
\newblock In {\em Advances in Neural Information Processing Systems}, volume~4, 1991.

\bibitem[CMA94]{connor1994}
J.~Connor, R.~Martin, and L.~Atlas.
\newblock Recurrent neural networks and robust time series prediction.
\newblock {\em IEEE Transactions on Neural Networks}, 5(2):240--254, 1994.

\bibitem[CMBB14]{cho2014properties}
K.~Cho, B.~Van Merri{\"e}nboer, D.~Bahdanau, and Y.~Bengio.
\newblock On the properties of neural machine translation: Encoder-decoder approaches.
\newblock {\em arXiv preprint arXiv:1409.1259}, 2014.

\bibitem[Elm90]{ELMAN1990179}
J.~Elman.
\newblock Finding structure in time.
\newblock {\em Cognitive Science}, 14(2):179--211, 1990.

\bibitem[GA78]{granger1978invertibility}
C.~Granger and A.~Andersen.
\newblock On the invertibility of time series models.
\newblock {\em Stochastic Processes and their Applications}, 8(1):87--92, 1978.

\bibitem[GBC16]{GoodBengCour16}
I.~Goodfellow, Y.~Bengio, and A.~Courville.
\newblock {\em Deep Learning}.
\newblock MIT Press, Cambridge, MA, USA, 2016.

\bibitem[HK08]{autoarima}
R.~Hyndman and Y.~Khandakar.
\newblock Automatic time series forecasting: The forecast package for {R}.
\newblock {\em Journal of Statistical Software}, 26(3):1--22, 2008.

\bibitem[HS97]{Hochreiter.1997}
S.~Hochreiter and J.~Schmidhuber.
\newblock Long short-term memory.
\newblock {\em Neural Computation}, 9(8):1735--1780, 1997.

\bibitem[HZL{\etalchar{+}}19]{han2019review}
Z.~Han, J.~Zhao, H.~Leung, K.~Ma, and W.~Wang.
\newblock A review of deep learning models for time series prediction.
\newblock {\em IEEE Sensors Journal}, 21(6):7833--7848, 2019.

\bibitem[IKB21]{immer2021improving}
A.~Immer, M.~Korzepa, and M.~Bauer.
\newblock Improving predictions of bayesian neural nets via local linearization.
\newblock In {\em International Conference on Artificial Intelligence and Statistics}, pages 703--711. PMLR, 2021.

\bibitem[Jor86]{jordan:attractor}
M.~Jordan.
\newblock Attractor dynamics and parallelism in a connectionist sequential machine.
\newblock In {\em Proceedings of the Eighth Annual Conference of the Cognitive Science Society}, pages 531--546, Hillsdale, NJ: Erlbaum, 1986.

\bibitem[KB14]{kingma2014method}
D.~Kingma and J.~Ba.
\newblock Adam: A method for stochastic optimization, 2014.
\newblock Published as a conference paper at the 3rd International Conference for Learning Representations, San Diego, 2015.

\bibitem[LWG93]{lee1993testing}
T.-H. Lee, H.~White, and C.~Granger.
\newblock Testing for neglected nonlinearity in time series models: A comparison of neural network methods and alternative tests.
\newblock {\em Journal of Econometrics}, 56(3):269--290, 1993.

\bibitem[MSA18]{Makridakis.2018}
S.~Makridakis, E.~Spiliotis, and V.~Assimakopoulos.
\newblock The m4 competition: Results, findings, conclusion and way forward.
\newblock {\em International Journal of Forecasting}, 34(4):802--808, 2018.

\bibitem[Rao81]{rao1981theory}
T.~Rao.
\newblock On the theory of bilinear time series models.
\newblock {\em Journal of the Royal Statistical Society: Series B (Methodological)}, 43(2):244--255, 1981.

\bibitem[RP96]{rivals1996black}
I.~Rivals and L.~Personnaz.
\newblock Black-box modeling with state-space neural networks.
\newblock In {\em Neural Adaptive Control Technology}, pages 237--264. World Scientific, 1996.

\bibitem[Sax97]{saxen1997equivalence}
H.~Sax{\'e}n.
\newblock On the equivalence between {ARMA} models and simple recurrent neural networks.
\newblock In {\em Applications of Computer Aided Time Series Modeling}, pages 281--289. Springer, 1997.

\bibitem[SCW{\etalchar{+}}15]{NIPS2015_07563a3f}
X.~Shi, Z.~Chen, H.~Wang, D.-Y. Yeung, W.-K. Wong, and W.-C. Woo.
\newblock Convolutional {LSTM} network: A machine learning approach for precipitation nowcasting.
\newblock In C.~Cortes, N.~Lawrence, D.~Lee, M.~Sugiyama, and R.~Garnett, editors, {\em Advances in Neural Information Processing Systems}, volume~28. Curran Associates, Inc., 2015.

\bibitem[SFGJ20]{Salinas.2020}
D.~Salinas, V.~Flunkert, J.~Gasthaus, and T.~Januschowski.
\newblock Deepar: Probabilistic forecasting with autoregressive recurrent networks.
\newblock {\em International Journal of Forecasting}, 36(3):1181--1191, 2020.

\bibitem[SGL{\etalchar{+}}17]{shi2017deep}
X.~Shi, Z.~Gao, L.~Lausen, H.~Wang, D.-Y. Yeung, W.~K. Wong, and W.~C. Woo.
\newblock Deep learning for precipitation nowcasting: A benchmark and a new model, 2017.

\bibitem[SP10]{seabold2010statsmodels}
S.~Seabold and J.~Perktold.
\newblock statsmodels: Econometric and statistical modeling with python.
\newblock In {\em 9th Python in Science Conference}, 2010.

\bibitem[TB81]{tiao1981}
G.~Tiao and G.~Box.
\newblock Modeling multiple time series with applications.
\newblock {\em Journal of the American Statistical Association}, 76(376):802--816, 1981.

\bibitem[TL09]{tong2009threshold}
H.~Tong and K.~Lim.
\newblock Threshold autoregression, limit cycles and cyclical data.
\newblock In {\em Exploration Of A Nonlinear World: An Appreciation of Howell Tong's Contributions to Statistics}, pages 9--56. World Scientific, 2009.

\bibitem[TLY{\etalchar{+}}19]{tian2019generative}
L.~Tian, X.~Li, Y.~Ye, P.~Xie, and Y.~Li.
\newblock A generative adversarial gated recurrent unit model for precipitation nowcasting.
\newblock {\em IEEE Geoscience and Remote Sensing Letters}, 17(4):601--605, 2019.

\bibitem[Tri15]{electicitydata}
A.~Trindade.
\newblock Electricityloaddiagrams20112014, 2015.

\bibitem[YRD16]{yu.2016}
H-F. Yu, N.~Rao, and I.~Dhillon.
\newblock Temporal regularized matrix factorization for high-dimensional time series prediction.
\newblock In {\em NIPS}, pages 847--855, 2016.

\bibitem[Zha03]{zhang2003time}
G.~Zhang.
\newblock Time series forecasting using a hybrid {ARIMA} and neural network model.
\newblock {\em Neurocomputing}, 50:159--175, 2003.

\end{thebibliography}
}

\clearpage
\appendix

\section{Derivation of ARMA reparametrization} \label{app:derivation}

This shows how to rewrite the ARMA model. We start with
\begin{equation*}
    x_t = \alpha + \sum_{i=1}^p \beta_i x_{t-i} + \sum_{j=1}^q \gamma_j \varepsilon_{t-j} + \varepsilon_t
\end{equation*}
and use the definition of $\hat{x}_t \coloneqq x_t-\varepsilon_t$ to get
\begin{equation*}
    \hat{x}_t = \alpha + \sum_{i=1}^p \beta_i x_{t-i} + \sum_{j=1}^q \gamma_j \varepsilon_{t-j}.
\end{equation*}
We now replace each $\varepsilon_{t-j}$ with $x_{t-j} - \hat{x}_{t-j}$
\begin{equation*}
    \hat{x}_t = \alpha + \sum_{i=1}^p \beta_i x_{t-i} + \sum_{j=1}^q \gamma_j(x_{t-j} - \hat{x}_{t-j}) = \alpha +
    \sum_{i=1}^p \beta_i x_{t-i} + \sum_{i=1}^q \gamma_i x_{t-i} - \sum_{j=1}^q  \gamma_j \hat{x}_{t-j}.
\end{equation*}
We see that for all indices $i \leq \min(p,q)$ the common factor of $x_{t-i}$ is
$\beta_i + \gamma_i$, if $p>q$ and $i>q$ the factor is $\beta_i$ and if $q>p$
and $i>p$ then the factor is $\gamma_i$, yielding the desired result.

\section{Additional experimental details}

\subsection{ARMA parameter recovery} \label{app:paramrecov}
\begin{figure}[!h]
    \centering
    \includegraphics[width=0.7\textwidth]{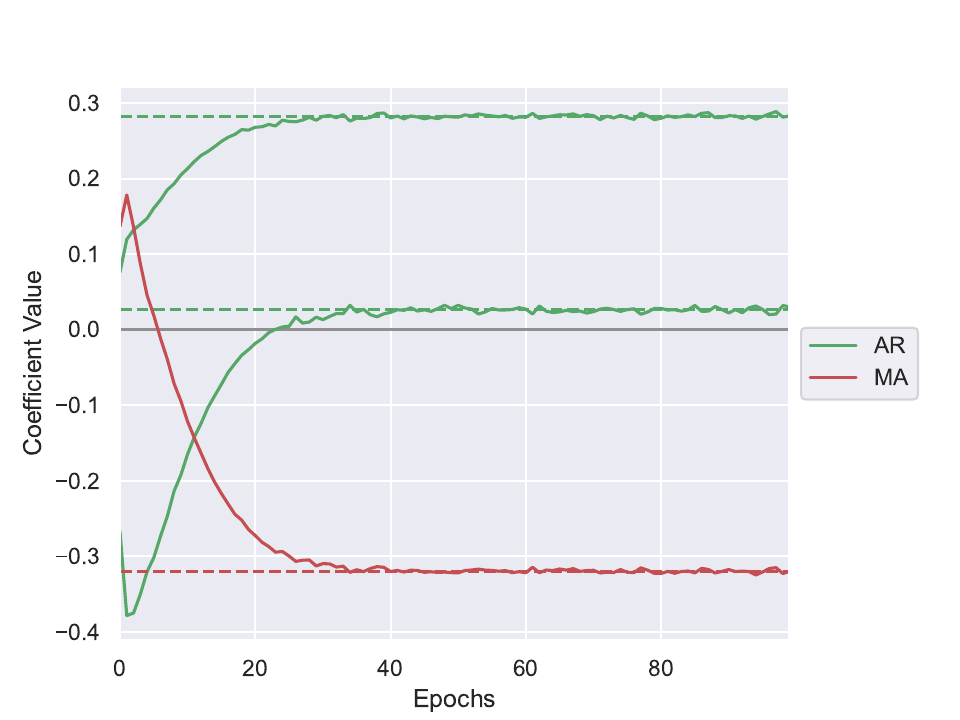}
    \caption{Exemplary optimization paths for a single linear ARMA($2$,$1$) cell
        using stochastic gradient descent. After around 30 iterations, the models
        converge to the maximum likelihood coefficients.}
    \label{fig:arma_param_recovery}
\end{figure}

In order to investigate if the implemented cell recovers parameters of an
arbitrary ARMA model with coefficients estimated in a standard ARMA software
\cite{seabold2010statsmodels}, we simulate (V)ARMA processes for $25,000$ time
steps and all possible combinations of $p,q\in\{0,1,\ldots,5\}$. We then train a
neural network defined by a single linear ARMA cell on the data and check the
convergence against the values obtained by maximum likelihood estimation.
Results confirm that the ARMA cell can recover the coefficients for different
values of $p$ and $q$, and also in the multivariate setting.
Figure~\ref{fig:arma_param_recovery} visualizes one exemplary learning process.

\subsection{Elman parameter recovery} \label{app:paramrecov2}

\begin{figure}[!h]
    \centering
    \includegraphics[width=0.7\textwidth]{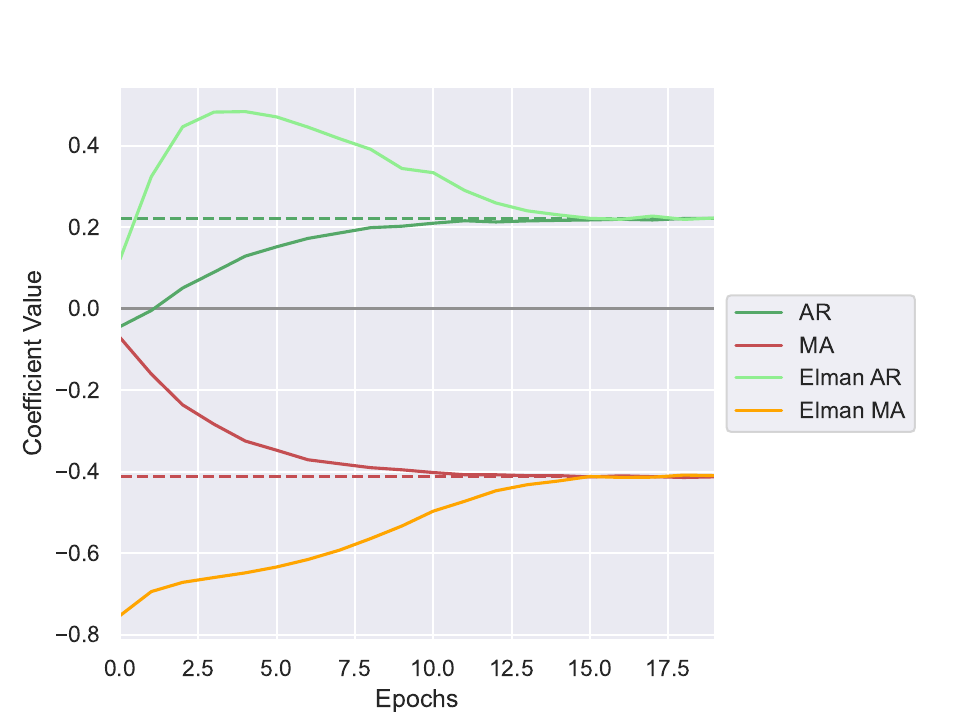}
    \caption{The ARMA cell and the Elman model converging to the same coefficients for an ARMA($1$,$1$) process.}
    \label{fig:arma_simple_param_recovery}
\end{figure}

We now demonstrate the equivalence of a network based on the ARMA cell and an
Elman network when only one MA lag is considered, \ie, when the ARMA cell is
restricted to $q=1$. For this, we take the ARMA(1,1) time series process used
also in our simulation studies and fit linearly activated single-unit models
based on both the ARMA cell and the Elman network. As shown in
Figure~\ref{fig:arma_simple_param_recovery}, both models converge to the same
parameters when applying the ARMA coefficient reparametrization as in (3).

\subsection{Simulation study}

\paragraph{Comparison between classical and first-order optimization.} In the
case where the data generating process is in fact a (V)ARMA process, we expect
the classical (V)ARMA model and the ARMA cell to perform similarly, but note
that the optimization using stochastic gradient descent can sometimes yield
better estimations of this process and hence outperform these classical models
despite having the exact same hypothesis space.

We start with a variety of synthetic data examples using time series models
defined in~\cite{lee1993testing}. Simulations include linear and non-linear, as
well as uni- and multivariate time series. All time series are of length 1000
and split into 70\% train and 30\% test data. The data generating processes
follow~\cite{lee1993testing} and include an ARMA process (ARMA), a threshold
autoregressive model (TAR), an autoregressive time series which is transformed
using the sign operation (SGN), a non-linear autoregressive series (NAR), a
heteroscedastic MA process (Heteroscedastic), a vector ARMA (VARMA), a
non-linear multivariate time series with quadratic lag structure (SQ), and an
exponential multivariate autoregressive time series (EXP). The exact
specification of the data generating processes are given below.

\paragraph{Description of simulated data generating processes.}

All error terms are a Gaussian white noise $\varepsilon_t \sim
    \mathcal{N}(0,1)$. The data generating processes were defined as follows:

\begin{itemize}
    \item ARMA(2,1)
          \begin{equation*}
              x_t = 0.1 x_{t-1} + 0.3 x_{t-2} - 0.4 \varepsilon_{t-1} + \varepsilon_t
          \end{equation*}
    \item Threshold autoregressive (TAR)
          \begin{equation*}
              x_t =
              \begin{cases}
                  \phantom{-}0.9x_{t-1} + \varepsilon_t & \mbox{ for } \lvert x_{t-1} \rvert \leq 1, \\
                  - 0.3x_{t-1} + \varepsilon_t          & \mbox{ for } \lvert x_{t-1} \rvert > 1
              \end{cases}
          \end{equation*}
    \item Sign autoregressive (SGN)
          \begin{equation*}
              x_t = sgn(x_{t-1}) + \varepsilon_t,
          \end{equation*}
          with
          \begin{equation*}
              sgn(x) =
              \begin{cases}
                  \phantom{-}1 \mbox{ for } x > 0, \\
                  \phantom{-}0 \mbox{ for } x = 0, \\
                  -1 \mbox{ for } x < 0
              \end{cases}
          \end{equation*}
    \item Non-linear autoregressive (NAR)
          \begin{equation*}
              x_t = \frac{0.7\lvert x_{t-1}\rvert}{\lvert x_{t-1}+2\rvert} + \varepsilon_t
          \end{equation*}
    \item Heteroskedastic MA(2)
          \begin{equation*}
              x_t = \varepsilon_{t} - 0.4\varepsilon_{t-1}+0.3\varepsilon_{t-2}+0.5\varepsilon_{t}\varepsilon_{t-2}
          \end{equation*}
    \item VARMA
          \begin{equation*}
              \begin{bmatrix}
                  x_{t,1} \\
                  x_{t,2}
              \end{bmatrix} =
              \begin{bmatrix}
                  0.1  & -0.2 \\
                  -0.2 & 0.1
              \end{bmatrix}
              \begin{bmatrix}
                  x_{t-1,1} \\
                  x_{t-1,2}
              \end{bmatrix} +
              \begin{bmatrix}
                  -0.4 & 0.2  \\
                  0.2  & -0.4
              \end{bmatrix}
              \begin{bmatrix}
                  \varepsilon_{t-1,1} \\
                  \varepsilon_{t-1,2}
              \end{bmatrix} +
              \begin{bmatrix}
                  \varepsilon_{t,1} \\
                  \varepsilon_{t,2}
              \end{bmatrix}
          \end{equation*}
    \item Square multivariate (SQ)
          \begin{align*}
              x_{t,1} & = 0.6x_{t-1} + \varepsilon_{t,1} \\
              x_{t,2} & = x_{t,1}^2 + \varepsilon_{t,2}
          \end{align*}
    \item Exponential multivariate (EXP)
          \begin{align*}
              x_{t,1} & = 0.6x_{t-1} + \varepsilon_{t,1}    \\
              x_{t,2} & = \exp(x_{t,1}) + \varepsilon_{t,2}
          \end{align*}

\end{itemize}

For the multivariate time series (VARMA, SQ, EXP), the second index of
$x_{t,i}$, $i \in \{1,2\}$, refers to the individual components.

\begin{table}[t]
    \caption{\small Comparisons of different methods (rows) and different data
        generating processes (columns) for univariate time series using the average
        RMSE $\pm$ the standard deviation of 30 independent runs. The best
        performing method is highlighted in bold, the second-best in italics.}
    \label{simulation_1}
    \begin{center}
        \begin{small}
            \begin{sc}
                \begin{tabular}{llllll}
                    \toprule
                    {}                   & ARMA                    & TAR
                                         & SGN                     & NAR                    & Heteroskedastic \\
                    model                &                         &
                                         &                         &                        &                 \\
                    \midrule
                    $\text{ARMA}$        & 2.04$\pm$0.35           &
                    2.92$\pm$3.20        & 2.36$\pm$2.67           & 2.67$\pm$4.92
                                         & 1.19$\pm$0.15                                                      \\
                    $\text{ShallowARMA}$ & \bfseries 1.96$\pm$0.09 & \bfseries
                    1.09$\pm$0.12        & 1.10$\pm$0.07           & \itshape 1.02$\pm$0.05 &
                    \bfseries 1.11$\pm$0.06                                                                   \\
                    $\text{DeepARMA}$    & 1.97$\pm$0.10           & \itshape
                    1.24$\pm$0.47        & \bfseries 1.05$\pm$0.06 & 1.02$\pm$0.04          &
                    \itshape 1.11$\pm$0.06                                                                    \\
                    $\text{LSTM}$        & 1.98$\pm$0.10           &
                    1.38$\pm$0.42        & 1.19$\pm$0.16           & 1.02$\pm$0.04
                                         & 1.15$\pm$0.09                                                      \\
                    $\text{DeepLSTM}$    & 2.02$\pm$0.10           &
                    1.46$\pm$0.55        & 1.17$\pm$0.13           & 1.02$\pm$0.04
                                         & 1.16$\pm$0.08                                                      \\
                    $\text{GRU}$         & \itshape 1.96$\pm$0.10  &
                    1.28$\pm$0.31        & \itshape 1.09$\pm$0.08  & \bfseries
                    1.02$\pm$0.04        & 1.13$\pm$0.07                                                      \\
                    $\text{DeepGRU}$     & 1.99$\pm$0.09           &
                    1.24$\pm$0.36        & 1.09$\pm$0.12           & 1.02$\pm$0.04
                                         & 1.12$\pm$0.06                                                      \\
                    $\text{Simple}$      & 1.99$\pm$0.09           &
                    1.29$\pm$0.31        & 1.14$\pm$0.09           & 1.04$\pm$0.04
                                         & 1.13$\pm$0.08                                                      \\
                    $\text{DeepSimple}$  & 2.01$\pm$0.11           &
                    1.47$\pm$0.57        & 1.15$\pm$0.10           & 1.03$\pm$0.04
                                         & 1.16$\pm$0.10                                                      \\
                    \bottomrule
                \end{tabular}
            \end{sc}
        \end{small}
    \end{center}
    \vskip -0.1in
\end{table}

\paragraph{Results.} The results in Table~\ref{simulation_1} suggest that the
ShallowARMA approach emulating an ARMA model in a neural network works well for
all linear- and non-linear datasets. In terms of robustness, the lower RMSE and
high standard deviation of the ARMA model on the ARMA process shows that fitting
an ARMA model in a neural network with stochastic gradient descent can, in fact,
be more robust than the standard software~\cite{autoarima,
    seabold2010statsmodels}. While the classical ARMA did match the performance of
its neural counterpart it some cases, the average RMSE is worse, as it did not
converge in all runs, even for the linear time series. The performance of the
DeepARMA approach is slightly worse compared to the ShallowARMA in most cases.
The performance of LSTM, GRU, and the simple RNN are all similar, with all
methods matching the ARMA cells in some cases, and falling slightly behind in
others. As expected, the classical ARMA approach does not work well for
non-linear data generating processes (TAR, SGN, NAR) and yields unstable
predictions underpinned by the large standard deviations in RMSE values.

\begin{table}[t]
    \caption{\small Comparisons of different methods (rows) and different data
        generating processes (columns) for multivariate time series using the
        average RMSE $\pm$ the standard deviation of 30 independent runs. The best
        performing method is highlighted in bold, the second-best in italics.}
    \label{simulation_2}
    \begin{center}
        \begin{small}
            \begin{sc}
                \begin{tabular}{lccc}
                    {}                   & VARMA                   & EXP       & SQ
                    \\
                    \midrule
                    $\text{VARMA}$       & \bfseries 1.00$\pm$0.03 &
                    3.35$\pm$0.69        & 1.90$\pm$0.14                            \\
                    $\text{ShallowARMA}$ & \itshape 1.00$\pm$0.03  & \itshape
                    3.14$\pm$0.74        & \bfseries 1.75$\pm$0.12                  \\
                    $\text{DeepARMA}$    & 1.01$\pm$0.03           & \bfseries
                    3.10$\pm$0.75        & \itshape 1.76$\pm$0.11                   \\
                    $\text{LSTM}$        & 1.01$\pm$0.04           &
                    3.26$\pm$0.73        & 1.83$\pm$0.16                            \\
                    $\text{DeepLSTM}$    & 1.03$\pm$0.04           &
                    3.35$\pm$0.68        & 1.86$\pm$0.14                            \\
                    $\text{GRU}$         & 1.02$\pm$0.04           &
                    3.19$\pm$0.75        & 1.80$\pm$0.13                            \\
                    $\text{DeepGRU}$     & 1.01$\pm$0.04           &
                    3.22$\pm$0.80        & 1.82$\pm$0.18                            \\
                    $\text{SIMPLE}$      & 1.02$\pm$0.03           &
                    3.29$\pm$0.70        & 1.80$\pm$0.12                            \\
                    $\text{DeepSimple}$  & 1.02$\pm$0.03           &
                    3.30$\pm$0.84        & 1.83$\pm$0.13                            \\
                    \bottomrule
                \end{tabular}
            \end{sc}
        \end{small}
    \end{center}
    \vskip -0.1in
\end{table}

For multivariate time series results of the simulation are summarized in
Table~\ref{simulation_2}. The results again show that the ShallowARMA model
matches the performance of the classical VARMA model for a dataset that is also
based on a VARMA process. For other types of data generation, the ShallowARMA
model and DeepARMA model work similarly well. Both outperform the other neural
cells, which in turn yield better results than the VARMA baseline.

In summary, findings suggest that ARMA cells work well for simpler linear and
non-linear data generating processes while being much more stable than a
classical ARMA approach.

\subsection{Ablation studies} \label{app:ablation}

To explore the validity of our presented results, we perform a series of
ablation studies. Two important influence factors on the performance of time
series models are the length of the time series and the forecasting horizon,
which we subsequently assess in the controlled setting of our simulation study.

\subsubsection{Time series length}
In order to investigate the influence of the time series length on the
performance reported in previous simulations, we vary $T\in\{200, 1000, 10000\}$
and re-run the experiments reported in Table~\ref{simulation_1}
and~\ref{simulation_2}. The results are given in Tables~\ref{tab:uni_extra}
and~\ref{tab:multi_extra}. In summary, the rank of the different methods is
similar to the aforementioned results, and ShallowARMA yields the best results
in most settings. There is, however, a clear trend in that the performance
differences between the different cell types become irrelevant for an increase
in $T$. For example, for the multivariate time series study setting SQ, the
ShallowARMA yields a notably better MSE for $T=200$ compared to the DeepSimple
cell ($1.97\pm 0.41$ vs.~$3.00\pm 3.95$), the performances are almost identical
for $T=10,000$ observations ($1.78\pm 0.05$ vs.~$1.80 \pm 0.07$).

\begin{table}[!h]
    \begin{center}
        \begin{subtable}[h]{0.8\textwidth}
            \centering
            \begin{tabular}{llllll}
                \toprule
                {}                   & ARMA                    & TAR
                                     & SGN                     & NAR       &
                Hetero                                                       \\
                \midrule
                $\text{ARMA}$        & 2.19$\pm$0.97           &
                1.77$\pm$0.86        & 1.80$\pm$1.54           &
                1.04$\pm$0.15        & 1.38$\pm$0.56                         \\
                $\text{ShallowARMA}$ & \bfseries 2.00$\pm$0.21 & \bfseries
                1.51$\pm$0.46        & \bfseries 1.29$\pm$0.10 & \itshape
                1.03$\pm$0.13        & \bfseries 1.17$\pm$0.15               \\
                $\text{DeepARMA}$    & \itshape 2.04$\pm$0.22  &
                1.71$\pm$0.43        & \itshape 1.32$\pm$0.16  &
                1.03$\pm$0.13        & \itshape 1.21$\pm$0.17                \\
                $\text{LSTM}$        & 2.08$\pm$0.24           &
                2.05$\pm$0.98        & 1.40$\pm$0.16           &
                1.03$\pm$0.14        & 1.22$\pm$0.17                         \\
                $\text{DeepLSTM}$    & 2.09$\pm$0.24           &
                2.03$\pm$0.51        & 1.46$\pm$0.17           &
                1.03$\pm$0.13        & 1.24$\pm$0.17                         \\
                $\text{GRU}$         & 2.08$\pm$0.24           & \itshape
                1.71$\pm$0.49        & 1.34$\pm$0.16           &
                1.03$\pm$0.13        & 1.23$\pm$0.17                         \\
                $\text{DeepGRU}$     & 2.08$\pm$0.22           &
                1.77$\pm$0.43        & 1.38$\pm$0.15           & \bfseries
                1.03$\pm$0.13        & 1.23$\pm$0.18                         \\
                $\text{SIMPLE}$      & 2.32$\pm$0.69           &
                2.14$\pm$0.91        & 1.43$\pm$0.22           &
                1.16$\pm$0.53        & 1.34$\pm$0.36                         \\
                $\text{DeepSIMPLE}$  & 2.30$\pm$0.87           &
                2.34$\pm$1.47        & 1.70$\pm$0.81           &
                1.10$\pm$0.29        & 1.25$\pm$0.18                         \\
                \bottomrule
            \end{tabular}
            \caption{200 observations}
        \end{subtable}
        \hfill
        \begin{subtable}[h]{0.8\textwidth}
            \centering
            \begin{tabular}{llllll}
                \toprule
                {}                   & ARMA                    & TAR
                                     & SGN                     & NAR       &
                Hetero                                                       \\
                \midrule
                $\text{ARMA}$        & 2.07$\pm$0.29           &
                1.98$\pm$1.98        & 1.71$\pm$1.80           &
                1.94$\pm$3.46        & 1.48$\pm$1.60                         \\
                $\text{ShallowARMA}$ & \bfseries 2.02$\pm$0.11 & \bfseries
                1.12$\pm$0.14        & 1.12$\pm$0.05           & \bfseries
                1.00$\pm$0.05        & \bfseries 1.11$\pm$0.06               \\
                $\text{DeepARMA}$    & \itshape 2.04$\pm$0.11  & \itshape
                1.18$\pm$0.29        & \bfseries 1.07$\pm$0.05 & \itshape
                1.00$\pm$0.05        & \itshape 1.12$\pm$0.06                \\
                $\text{LSTM}$        & 2.06$\pm$0.12           &
                1.22$\pm$0.25        & 1.16$\pm$0.10           &
                1.00$\pm$0.05        & 1.15$\pm$0.06                         \\
                $\text{DeepLSTM}$    & 2.11$\pm$0.14           &
                1.18$\pm$0.20        & 1.16$\pm$0.11           &
                1.00$\pm$0.05        & 1.17$\pm$0.09                         \\
                $\text{GRU}$         & 2.06$\pm$0.13           &
                1.29$\pm$0.30        & 1.14$\pm$0.09           &
                1.00$\pm$0.05        & 1.13$\pm$0.06                         \\
                $\text{DeepGRU}$     & 2.08$\pm$0.13           &
                1.21$\pm$0.26        & \itshape 1.11$\pm$0.09  &
                1.00$\pm$0.05        & 1.13$\pm$0.07                         \\
                $\text{SIMPLE}$      & 2.07$\pm$0.13           &
                1.31$\pm$0.25        & 1.18$\pm$0.11           &
                1.01$\pm$0.05        & 1.15$\pm$0.08                         \\
                $\text{DeepSIMPLE}$  & 2.09$\pm$0.12           &
                1.43$\pm$0.45        & 1.16$\pm$0.11           &
                1.01$\pm$0.06        & 1.15$\pm$0.08                         \\
                \bottomrule
            \end{tabular}
            \caption{1000 observations}
        \end{subtable}
        \hfill
        \begin{subtable}[h]{0.8\textwidth}
            \centering
            \begin{tabular}{llllll}
                \toprule
                {}                   & ARMA                    & TAR
                                     & SGN                     & NAR       &
                Hetero                                                       \\
                \midrule
                $\text{ARMA}$        & 2.04$\pm$0.23           &
                1.60$\pm$0.94        & 1.94$\pm$3.10           &
                1.05$\pm$0.14        & 1.16$\pm$0.11                         \\
                $\text{ShallowARMA}$ & \bfseries 2.00$\pm$0.03 & \bfseries
                1.01$\pm$0.01        & 1.06$\pm$0.03           & \bfseries
                1.00$\pm$0.01        & 1.06$\pm$0.02                         \\
                $\text{DeepARMA}$    & 2.00$\pm$0.04           &
                1.06$\pm$0.23        & \bfseries 1.02$\pm$0.03 & \itshape
                1.00$\pm$0.01        & 1.05$\pm$0.02                         \\
                $\text{LSTM}$        & \itshape 2.00$\pm$0.03  &
                1.06$\pm$0.23        & \itshape 1.03$\pm$0.01  &
                1.00$\pm$0.01        & 1.04$\pm$0.02                         \\
                $\text{DeepLSTM}$    & 2.02$\pm$0.06           &
                1.11$\pm$0.33        & 1.06$\pm$0.12           &
                1.01$\pm$0.01        & 1.04$\pm$0.04                         \\
                $\text{GRU}$         & 2.01$\pm$0.04           &
                1.19$\pm$0.45        & 1.07$\pm$0.12           &
                1.00$\pm$0.01        & \itshape 1.04$\pm$0.02                \\
                $\text{DeepGRU}$     & 2.01$\pm$0.04           &
                1.19$\pm$0.44        & 1.04$\pm$0.10           &
                1.00$\pm$0.01        & \bfseries 1.03$\pm$0.02               \\
                $\text{SIMPLE}$      & 2.02$\pm$0.04           & \itshape
                1.02$\pm$0.02        & 1.08$\pm$0.08           &
                1.00$\pm$0.01        & 1.06$\pm$0.02                         \\
                $\text{DeepSIMPLE}$  & 2.02$\pm$0.06           &
                1.15$\pm$0.37        & 1.04$\pm$0.08           &
                1.00$\pm$0.01        & 1.05$\pm$0.02                         \\
                \bottomrule
            \end{tabular}
            \caption{10000 observations}
        \end{subtable}
        \caption{Comparisons of different methods (rows) and different data
            generating processes (columns) across different time series lengths
            (200 (a), 1000 (b), 10000 (c)) for univariate time series using the
            average RMSE $\pm$ the standard deviation of 30 independent runs. The
            best performing method is highlighted in bold, the second-best in
            italics.}
        \label{tab:uni_extra}
    \end{center}
\end{table}

\begin{table}[!h]
    \begin{center}
        \begin{subtable}[h]{0.8\textwidth}
            \centering
            \begin{tabular}{llll}
                \toprule
                {}                   & VARMA 200               &
                EXP 200              & SQ 200                              \\
                $\text{VARMA}$       & 1.02$\pm$0.08           &
                3.11$\pm$1.67        & 2.07$\pm$0.44                       \\
                $\text{ShallowARMA}$ & \bfseries 1.01$\pm$0.06 & \bfseries
                2.97$\pm$1.68        & \bfseries 1.97$\pm$0.41             \\
                $\text{DeepARMA}$    & 1.02$\pm$0.07           & \itshape
                3.02$\pm$1.68        & \itshape 2.04$\pm$0.47              \\
                $\text{LSTM}$        & \itshape 1.01$\pm$0.06  &
                3.24$\pm$1.77        & 2.17$\pm$0.57                       \\
                $\text{DeepLSTM}$    & 1.02$\pm$0.06           &
                3.12$\pm$1.70        & 2.16$\pm$0.56                       \\
                $\text{GRU}$         & 1.02$\pm$0.07           &
                3.12$\pm$1.72        & 2.06$\pm$0.47                       \\
                $\text{DeepGRU}$     & 1.02$\pm$0.06           &
                3.15$\pm$1.73        & 2.14$\pm$0.55                       \\
                $\text{SIMPLE}$      & 1.09$\pm$0.13           &
                3.60$\pm$2.04        & 2.79$\pm$1.59                       \\
                $\text{DeepSIMPLE}$  & 1.06$\pm$0.09           &
                3.32$\pm$1.83        & 3.00$\pm$3.95                       \\
                \bottomrule
            \end{tabular}
            \caption{200 observations}
        \end{subtable}
        \hfill
        \begin{subtable}[h]{0.8\textwidth}
            \centering
            \begin{tabular}{llll}
                \toprule
                {}                   & VARMA 1k                &
                EXP 1k               & SQ 1k                               \\
                $\text{VARMA}$       & \bfseries 1.00$\pm$0.03 &
                2.92$\pm$0.61        & 1.88$\pm$0.17                       \\
                $\text{ShallowARMA}$ & \itshape 1.01$\pm$0.04  & \bfseries
                2.70$\pm$0.63        & \bfseries 1.75$\pm$0.16             \\
                $\text{DeepARMA}$    & 1.01$\pm$0.04           & \itshape
                2.71$\pm$0.63        & \itshape 1.75$\pm$0.15              \\
                $\text{LSTM}$        & 1.02$\pm$0.04           &
                2.80$\pm$0.65        & 1.81$\pm$0.17                       \\
                $\text{DeepLSTM}$    & 1.04$\pm$0.04           &
                2.86$\pm$0.63        & 1.84$\pm$0.17                       \\
                $\text{GRU}$         & 1.02$\pm$0.04           &
                2.76$\pm$0.61        & 1.78$\pm$0.17                       \\
                $\text{DeepGRU}$     & 1.02$\pm$0.04           &
                2.81$\pm$0.66        & 1.81$\pm$0.18                       \\
                $\text{SIMPLE}$      & 1.03$\pm$0.04           &
                3.18$\pm$1.67        & 1.82$\pm$0.19                       \\
                $\text{DeepSIMPLE}$  & 1.04$\pm$0.04           &
                2.89$\pm$0.72        & 1.85$\pm$0.18                       \\
                \bottomrule
            \end{tabular}
            \caption{1000 observations}
        \end{subtable}
        \hfill
        \begin{subtable}[h]{0.8\textwidth}
            \centering
            \begin{tabular}{llll}
                \toprule
                {}                   & VARMA 10k               &
                EXP 10k              & SQ 10k                              \\
                $\text{VARMA}$       & \bfseries 1.00$\pm$0.01 &
                3.04$\pm$0.39        & 1.94$\pm$0.06                       \\
                $\text{ShallowARMA}$ & \itshape 1.00$\pm$0.01  & \bfseries
                2.80$\pm$0.41        & \itshape 1.78$\pm$0.05              \\
                $\text{DeepARMA}$    & 1.00$\pm$0.01           & \itshape
                2.81$\pm$0.41        & \bfseries 1.78$\pm$0.05             \\
                $\text{LSTM}$        & 1.00$\pm$0.01           &
                2.84$\pm$0.46        & 1.79$\pm$0.06                       \\
                $\text{DeepLSTM}$    & 1.01$\pm$0.01           &
                2.90$\pm$0.46        & 1.81$\pm$0.07                       \\
                $\text{GRU}$         & 1.00$\pm$0.01           &
                2.82$\pm$0.41        & 1.79$\pm$0.06                       \\
                $\text{DeepGRU}$     & 1.00$\pm$0.01           &
                2.87$\pm$0.43        & 1.80$\pm$0.06                       \\
                $\text{SIMPLE}$      & 1.00$\pm$0.01           &
                2.83$\pm$0.42        & 1.79$\pm$0.05                       \\
                $\text{DeepSIMPLE}$  & 1.01$\pm$0.01           &
                2.86$\pm$0.41        & 1.80$\pm$0.07                       \\
                \bottomrule
            \end{tabular}
            \caption{10000 observations}
        \end{subtable}
        \caption{Comparisons of different methods (rows) and different data
            generating processes (columns) across different time series lengths (200
            (a), 1000 (b), 10000 (c)) for multivariate time series using the average
            RMSE $\pm$ the standard deviation of 30 independent runs. The best
            performing method is highlighted in bold, the second-best in italics.}
        \label{tab:multi_extra}
    \end{center}
\end{table}

\clearpage
\subsubsection{Forecasting horizon}

Similar to the previous ablation study, we reran the experiments but now alter
the forecasting horizon by comparing a one-step, 10-step, and 20-step forecast
for $T=1000$. The results can be found in Table~\ref{tab:horizon_extra}
and~\ref{tab:horizon_extra_multi}. In the univariate case for forecasting
horizons greater one, the different ARMA variations do not outperform other
approaches anymore and DeepLSTM, GRU, or Deep GRU yield the best results in many
cases. The performance values, however, are in most cases within one standard
deviation of those by the Shallow- or DeepARMA approach. For the multivariate
case, the classical VARMA model provides the best forecast for all multi-step
ahead forecast scenarios, closely followed by the Shallow- and DeepARMA models.

\begin{table}[!h]
    \begin{center}
        \begin{subtable}[h]{0.8\textwidth}
            \centering
            \begin{tabular}{llllll}
                \toprule
                {}                      & ARMA                    & TAR                     &
                SGN                     & NAR                     & Hetero                    \\
                \midrule
                $\text{ARMA}$           & 2.07$\pm$0.29           & 1.98$\pm$1.98           &
                1.71$\pm$1.80           & 1.94$\pm$3.46           & 1.48$\pm$1.60             \\
                $\text{ShallowARMA}$    & \bfseries 2.02$\pm$0.11 & \bfseries 1.12$\pm$0.14 &
                1.12$\pm$0.05           & \bfseries 1.00$\pm$0.05 & \bfseries 1.11$\pm$0.06   \\
                $\text{DeepARMA}$       & \itshape 2.04$\pm$0.11  & \itshape 1.18$\pm$0.29  &
                \bfseries 1.07$\pm$0.05 & \itshape 1.00$\pm$0.05  & \itshape 1.12$\pm$0.06
                \\
                $\text{LSTM}$           & 2.06$\pm$0.12           & 1.22$\pm$0.25           &
                1.16$\pm$0.10           & 1.00$\pm$0.05           & 1.15$\pm$0.06             \\
                $\text{DeepLSTM}$       & 2.11$\pm$0.14           & 1.18$\pm$0.20           &
                1.16$\pm$0.11           & 1.00$\pm$0.05           & 1.17$\pm$0.09             \\
                $\text{GRU}$            & 2.06$\pm$0.13           & 1.29$\pm$0.30           &
                1.14$\pm$0.09           & 1.00$\pm$0.05           & 1.13$\pm$0.06             \\
                $\text{DeepGRU}$        & 2.08$\pm$0.13           & 1.21$\pm$0.26           &
                \itshape 1.11$\pm$0.09  & 1.00$\pm$0.05           & 1.13$\pm$0.07             \\
                $\text{SIMPLE}$         & 2.07$\pm$0.13           & 1.31$\pm$0.25           &
                1.18$\pm$0.11           & 1.01$\pm$0.05           & 1.15$\pm$0.08             \\
                $\text{DeepSIMPLE}$     & 2.09$\pm$0.12           & 1.43$\pm$0.45           &
                1.16$\pm$0.11           & 1.01$\pm$0.06           & 1.15$\pm$0.08             \\
                \bottomrule
            \end{tabular}
            \caption{1 step ahead}
        \end{subtable}
        \hfill
        \begin{subtable}[h]{0.8\textwidth}
            \centering
            \begin{tabular}{llllll}
                \toprule
                {}                      & ARMA                    & TAR                     &
                SGN                     & NAR                     & Hetero                    \\
                \midrule
                $\text{ARMA}$           & \bfseries 2.12$\pm$0.06 & 2.58$\pm$0.64           &
                1.44$\pm$0.08           & 1.03$\pm$0.04           & 1.28$\pm$0.06             \\
                $\text{ShallowARMA}$    & 2.16$\pm$0.14           & 2.33$\pm$0.34           &
                1.41$\pm$0.04           & 1.00$\pm$0.05           & 1.24$\pm$0.08             \\
                $\text{DeepARMA}$       & 2.17$\pm$0.13           & \itshape 2.26$\pm$0.33  &
                \itshape 1.41$\pm$0.04  & \bfseries 1.00$\pm$0.05 & 1.24$\pm$0.08             \\
                $\text{LSTM}$           & 2.17$\pm$0.13           & 2.48$\pm$1.22           &
                1.41$\pm$0.04           & 1.00$\pm$0.05           & 1.24$\pm$0.08             \\
                $\text{DeepLSTM}$       & 2.16$\pm$0.13           & 2.30$\pm$0.37           &
                \bfseries 1.41$\pm$0.04 & 1.00$\pm$0.05           & \itshape 1.24$\pm$0.08
                \\
                $\text{GRU}$            & 2.17$\pm$0.13           & \bfseries 2.25$\pm$0.30 &
                1.41$\pm$0.05           & 1.00$\pm$0.05           & 1.24$\pm$0.08             \\
                $\text{DeepGRU}$        & \itshape 2.16$\pm$0.13  & 2.29$\pm$0.35           &
                1.41$\pm$0.04           & \itshape 1.00$\pm$0.05  & \bfseries 1.24$\pm$0.08   \\
                $\text{SIMPLE}$         & 2.18$\pm$0.14           & 2.28$\pm$0.28           &
                1.42$\pm$0.05           & 1.01$\pm$0.06           & 1.29$\pm$0.15             \\
                $\text{DeepSIMPLE}$     & 2.22$\pm$0.23           & 2.34$\pm$0.42           &
                1.43$\pm$0.07           & 1.02$\pm$0.06           & 1.25$\pm$0.08             \\
                \bottomrule
            \end{tabular}
            \caption{10 step ahead}
        \end{subtable}
        \hfill
        \begin{subtable}[h]{0.8\textwidth}
            \centering
            \begin{tabular}{llllll}
                \toprule
                {}                      & ARMA                    & TAR                     &
                SGN                     & NAR                     & Hetero                    \\
                \midrule
                $\text{ARMA}$           & \bfseries 2.10$\pm$0.07 & 3.25$\pm$1.38           &
                1.43$\pm$0.08           & 1.02$\pm$0.02           & 1.24$\pm$0.05             \\
                $\text{ShallowARMA}$    & 2.17$\pm$0.13           & 2.33$\pm$0.32           &
                \bfseries 1.42$\pm$0.05 & 1.00$\pm$0.04           & 1.24$\pm$0.09
                \\
                $\text{DeepARMA}$       & 2.17$\pm$0.13           & 2.32$\pm$0.33           &
                \itshape 1.42$\pm$0.05  & 1.00$\pm$0.04           & \itshape 1.23$\pm$0.09
                \\
                $\text{LSTM}$           & 2.17$\pm$0.13           & 2.59$\pm$1.45           &
                1.43$\pm$0.06           & 1.00$\pm$0.04           & 1.24$\pm$0.09             \\
                $\text{DeepLSTM}$       & \itshape 2.17$\pm$0.13  & 2.33$\pm$0.35           &
                1.43$\pm$0.05           & \bfseries 1.00$\pm$0.04 & \bfseries 1.23$\pm$0.09   \\
                $\text{GRU}$            & 2.17$\pm$0.13           & \bfseries 2.32$\pm$0.31 &
                1.44$\pm$0.06           & 1.01$\pm$0.04           & 1.24$\pm$0.09             \\
                $\text{DeepGRU}$        & 2.17$\pm$0.13           & \itshape 2.32$\pm$0.32  &
                1.43$\pm$0.06           & \itshape 1.00$\pm$0.04  & 1.24$\pm$0.09             \\
                $\text{SIMPLE}$         & 2.18$\pm$0.14           & 2.44$\pm$0.45           &
                1.44$\pm$0.06           & 1.01$\pm$0.04           & 1.26$\pm$0.10             \\
                $\text{DeepSIMPLE}$     & 2.18$\pm$0.13           & 2.42$\pm$0.43           &
                1.44$\pm$0.07           & 1.07$\pm$0.23           & 1.24$\pm$0.09             \\
                \bottomrule
            \end{tabular}
            \caption{20 step ahead}
        \end{subtable}
        \caption{Comparisons of different methods (rows) and different data
            generating processes (columns) across different forecasting horizons (1 (a),
            10 (b), 20 (c)) for univariate time series using the average RMSE $\pm$ the
            standard deviation of 30 independent runs. The best performing method is
            highlighted in bold, the second-best in italics.}
        \label{tab:horizon_extra}
    \end{center}
\end{table}

\begin{table}[!h]
    \begin{center}
        \begin{subtable}[h]{0.8\textwidth}
            \centering
            \begin{tabular}{llll}
                \toprule
                {}                   & VARMA                   & EXP                     &
                SQ                                                                         \\
                \midrule
                $\text{VARMA}$       & \bfseries 1.00$\pm$0.03 & 2.92$\pm$0.61           &
                1.88$\pm$0.17                                                              \\
                $\text{ShallowARMA}$ & \itshape 1.01$\pm$0.04  & \bfseries 2.70$\pm$0.63 &
                \bfseries 1.75$\pm$0.16                                                    \\
                $\text{DeepARMA}$    & 1.01$\pm$0.04           & \itshape 2.71$\pm$0.63  &
                \itshape 1.75$\pm$0.15                                                     \\
                $\text{LSTM}$        & 1.02$\pm$0.04           & 2.80$\pm$0.65           &
                1.81$\pm$0.17                                                              \\
                $\text{DeepLSTM}$    & 1.04$\pm$0.04           & 2.86$\pm$0.63           &
                1.84$\pm$0.17                                                              \\
                $\text{GRU}$         & 1.02$\pm$0.04           & 2.76$\pm$0.61           &
                1.78$\pm$0.17                                                              \\
                $\text{DeepGRU}$     & 1.02$\pm$0.04           & 2.81$\pm$0.66           &
                1.81$\pm$0.18                                                              \\
                $\text{SIMPLE}$      & 1.03$\pm$0.04           & 3.18$\pm$1.67           &
                1.82$\pm$0.19                                                              \\
                $\text{DeepSIMPLE}$  & 1.04$\pm$0.04           & 2.89$\pm$0.72           &
                1.85$\pm$0.18                                                              \\
                \bottomrule
            \end{tabular}
            \caption{1 step ahead}
        \end{subtable}
        \hfill
        \begin{subtable}[h]{0.8\textwidth}
            \centering
            \begin{tabular}{llll}
                \toprule
                {}                   & VARMA                   & EXP                     &
                SQ                                                                         \\
                \midrule
                $\text{VARMA}$       & \bfseries 1.05$\pm$0.03 & \bfseries 2.98$\pm$0.80 &
                \bfseries 1.89$\pm$0.19                                                    \\
                $\text{ShallowARMA}$ & 1.05$\pm$0.03           & 3.03$\pm$0.81           &
                \itshape 1.91$\pm$0.19                                                     \\
                $\text{DeepARMA}$    & 1.05$\pm$0.04           & \itshape 3.02$\pm$0.81  &
                1.91$\pm$0.20                                                              \\
                $\text{LSTM}$        & 1.05$\pm$0.04           & 3.08$\pm$0.83           &
                1.93$\pm$0.20                                                              \\
                $\text{DeepLSTM}$    & \itshape 1.05$\pm$0.04  & 3.06$\pm$0.81           &
                1.93$\pm$0.21                                                              \\
                $\text{GRU}$         & 1.05$\pm$0.03           & 3.06$\pm$0.82           &
                1.92$\pm$0.20                                                              \\
                $\text{DeepGRU}$     & 1.05$\pm$0.04           & 3.06$\pm$0.83           &
                1.91$\pm$0.19                                                              \\
                $\text{SIMPLE}$      & 1.06$\pm$0.04           & 3.09$\pm$0.83           &
                1.98$\pm$0.22                                                              \\
                $\text{DeepSIMPLE}$  & 1.06$\pm$0.04           & 3.19$\pm$1.01           &
                1.95$\pm$0.20                                                              \\
                \bottomrule
            \end{tabular}
            \caption{10 step ahead}
        \end{subtable}
        \hfill
        \begin{subtable}[h]{0.8\textwidth}
            \centering
            \begin{tabular}{llll}
                \toprule
                {}                   & VARMA                   & EXP                     &
                SQ                                                                         \\
                \midrule
                $\text{VARMA}$       & \bfseries 1.05$\pm$0.04 & \bfseries 2.88$\pm$0.88 &
                \bfseries 1.97$\pm$0.20                                                    \\
                $\text{ShallowARMA}$ & 1.06$\pm$0.04           & \itshape 2.91$\pm$0.90  &
                \itshape 2.00$\pm$0.20                                                     \\
                $\text{DeepARMA}$    & 1.05$\pm$0.04           & 2.91$\pm$0.89           &
                2.00$\pm$0.20                                                              \\
                $\text{LSTM}$        & 1.05$\pm$0.04           & 2.93$\pm$0.90           &
                2.04$\pm$0.27                                                              \\
                $\text{DeepLSTM}$    & \itshape 1.05$\pm$0.04  & 2.97$\pm$0.95           &
                2.01$\pm$0.20                                                              \\
                $\text{GRU}$         & 1.05$\pm$0.04           & 2.98$\pm$1.06           &
                2.01$\pm$0.20                                                              \\
                $\text{DeepGRU}$     & 1.05$\pm$0.04           & 2.93$\pm$0.88           &
                2.01$\pm$0.21                                                              \\
                $\text{SIMPLE}$      & 1.06$\pm$0.04           & 3.06$\pm$0.93           &
                2.09$\pm$0.27                                                              \\
                $\text{DeepSIMPLE}$  & 1.06$\pm$0.04           & 2.98$\pm$0.89           &
                2.11$\pm$0.50                                                              \\
                \bottomrule
            \end{tabular}
            \caption{20 step ahead}
        \end{subtable}
        \caption{Comparisons of different methods (rows) and different data
            generating processes (columns) across different forecasting horizons (1 (a),
            10,(b) 20 (c)) for multivariate time series using the average RMSE $\pm$ the
            standard deviation of 30 independent runs. The best performing method is
            highlighted in bold, the second-best in italics.}
        \label{tab:horizon_extra_multi}
    \end{center}
\end{table}

\subsection{Description of benchmark datasets} \label{app:benchmarkdesc}

\paragraph{M4.} Stemming from the Makridakis Competitions \cite{Makridakis.2018}
(see \url{https://en.wikipedia .org/wiki/Makridakis_Competitions} for more
information), the M4 dataset contains 414 time series of hourly data. Every time
series has a different starting point and a length of 748 hours. To allow for
multivariate prediction, we take a subset of ten times series starting at the
same time and ending at the same time. We further take differences with a period
of one and 24 hours to improve stationarity and reduce seasonal effects,
respectively.

\paragraph{Traffic.} The traffic dataset can be downloaded from
\url{https://archive.ics.uci.edu/ml/datasets/PEMS-SF}. It consists of 963 car
lane occupancy rates with values between 0 and 1 taken from freeways in the San
Francisco Bay Area. Time series start on the first of January 2008 and last
until March 30 2009 with an observation frequency of 10 minutes. To condense the
information, an hourly aggregation is used \cite{yu.2016}, yielding time series
of length 10,560. We use the first ten time series and observations until
'2008-06-22 23:00:00', yielding a total of 4,167 observations per lane. We
further apply seasonal differencing with a seasonal period of 24 hours and take
first differences to reduce non-stationary behavior.

\paragraph{Electricity.} The electricity dataset can be downloaded from
\url{https://archive.ics.uci.edu/ml/datasets/ElectricityLoadDiagrams20112014}.
The dataset consists of electricity consumption (kWh) time series of 370
customers~\cite{electicitydata}.
Values correspond to electricity usage in a frequency of 15 minutes. In our
benchmarks, we aggregate the values to hourly consumption (see also
\cite{yu.2016} for justification of this approach). We use a subset of ten
customers and a time range from '2014-01-01 00:00:00' to '2014-09-07 23:00:00',
yielding a total of 6,000 observations per customer . We further apply seasonal
differencing with a period of 24 hours to reduce seasonal effects and take the
first differences for stationarity reasons.




\subsection{Architectures and search space} \label{app:arch}

For uni- and multivariate time series, all neural networks contain one to two
RNN layers of the respective RNN cell, yielding the shallow and deep versions of
the models, respectively. The cells in each layer contain one to five units with
a rectified linear activation function. In the ShallowARMA model, one cell is
activated linearly as shown in Figure~\ref{fig:multi},
resembling a hybrid model. A final fully connected layer with linear activation
and appropriate output shape is used to match the dimensions of the time series.
The lag values $p$ and $q$ are chosen from the interval $[1,4]$. The loss
function of all models is the mean squared error function. For training, the
Adam~\cite{kingma2014method}
optimizer is used in combination with an early stopping callback to prevent
overfitting. For all other model properties, the default values are used. For
tensor-variate time series, batch normalization layers are added between the RNN
layers, and adaptive learning is added to improve convergence. Each layer
contains 64 filters, so a 2D convolution is added to reduce the number of
channels appropriately.

\subsection{Computational environment}

All experiments and benchmarks were carried out on an internal cluster. Uni- and
multivariate time series were trained on a server with 10 vCPUs, running on an
Intel(R) Xeon(R) Gold 6148 CPU @ 2.40GHz physical CPU and 48Gb allocated memory.
Tensor-variate time series were trained on a server with 16 vCPUS, running on an
Intel(R) Xeon(R) Gold 6226R CPU @ 2.90GHz, 32Gb allocated memory, and a Nvidia
GeForce RTX 2080 Ti (11Gb).

\subsection{Additional details on experimental setup}
We now give additional details about our experimental setup, clarifying the
preprocessing steps as well as the training routines.

\subsection{General setup and optimization}

Across all simulations and benchmarks, we use the Adam~\cite{kingma2014method}
optimizer with a learning rate of 1e-3, and momentum parameters $\beta_1=0.9$
and $\beta_2=0.999$. We use a batch size of 32, early stopping with patience 10
iterations, and run the model for a maximum of 100 epochs.

\subsection{Competitor architectures}

The following abbreviations are used for the methods we compare the ARMA cell
against:
\begin{itemize}
    \item LSTM/GRU/SIMPLE: A single LSTM/GRU/SIMPLE cell with ReLU activation
          function and a sigmoid recurrent activation function. The kernel, recurrent,
          and bias initializers are chosen to be uniform, orthogonal, and zero,
          respectively. The number of units is chosen via hyperparameter optimization
          in the range of one to five.
    \item DeepLSTM/DeepGRU/DeepSIMPLE: This setup stacks two of the cells in
          their non-deep counterparts, with the first layer set to return sequences.
\end{itemize}

\subsection{Input and out formats of RNNs} \label{app:inpform}

The input of the RNN cells are subsequences of the time series $\boldsymbol{X}$.
For a sequence length $s$, the input shape for a non-ARMA cell RNN is given by
$s \times k$. The input of the ARMA cell has an additional dimension for the
lagged inputs, \ie, $s \times k \times p$.

The non-ARMA cell RNN with $u$ units returns an $s \times u$ time series. In
order to allow the ARMA cell to be stacked, we have an additional dimension in
the output. Specified by the keyword argument \verb|return_lags|, we either
return $s \times (d*u) \times 1$ if \verb|False|, or $s \times (d*u) \times q$
if \verb|True|. The factor $d$ ensures that a single unit ARMA cell returns the
same number of dimensions as its input, allowing it to capture a classical VARMA
model.

\subsection{Specific setups}

\subsubsection{Simulation study}
For the simulated time series, no preprocessing was necessary, as the data
generating processes are all stationary and also of the same magnitude.

\subsubsection{Benchmarks}
For the real-world datasets, preprocessing was performed to improve the
convergence of all models. In particular, we first apply differencing to remove
trends in the time series. We define $\Delta_k$ to be the difference to the
$k$'th lag of the time series. For M4, traffic, and electricity, we use
$\Delta_1 \Delta_{24} \boldsymbol{X}$. We then standardize the resulting time
series with their empirical mean and variance.

\subsubsection{Integration with state-of-the-art forecasting frameworks}
\label{app:deepar}

For the DeepAR benchmark, we use a larger dataset by looking at the first 100
columns of the M4, traffic, and electricity datasets.

\paragraph{Architectures.}

We use the following two types of DeepAR architectures:

\begin{itemize}
    \item SINGLE: The SINGLE DeepAR model consists of one RNN block (either LSTM
          or ARMA) with $4 (1 + log(d))$ units for the LSTM and one unit for the ARMA
          cell, a dropout rate of 0.2, and returns a sequence that is further
          processed by a fully-connected layer with $s (1 + log(d))$ number of units
          where $s\in\{1,4\}$ is a hyperparameter and $\text{tanh}$ activation. The resulting
          output is then fed into a Gaussian distribution layer that multiplies the
          number of input units by two to define both a mean and standard deviation
          for all output units by multiplying the inputs with a weight matrix of
          respective size. For the ARMA cell-based DeepAR model, we have the
          additional hyperparameters $p$ and $q$, which we optimize over the range
          from 1 to 4, again only considering $p=q$ for computational reasons.
    \item STACKED: The STACKED model uses the same architecture as SINGLE but
          combines two of the RNN cells specified by the SINGLE model.
\end{itemize}

\subsection{Investigation of parameter influence}

In the following, we investigate the influence of the number of parameters on
the performance and provide a summary of the resulting hyperparameter
optimization results.

\paragraph{Relation between number of parameters and performance.}

\begin{figure}
    \centering
    \includegraphics[width=0.6\textwidth]{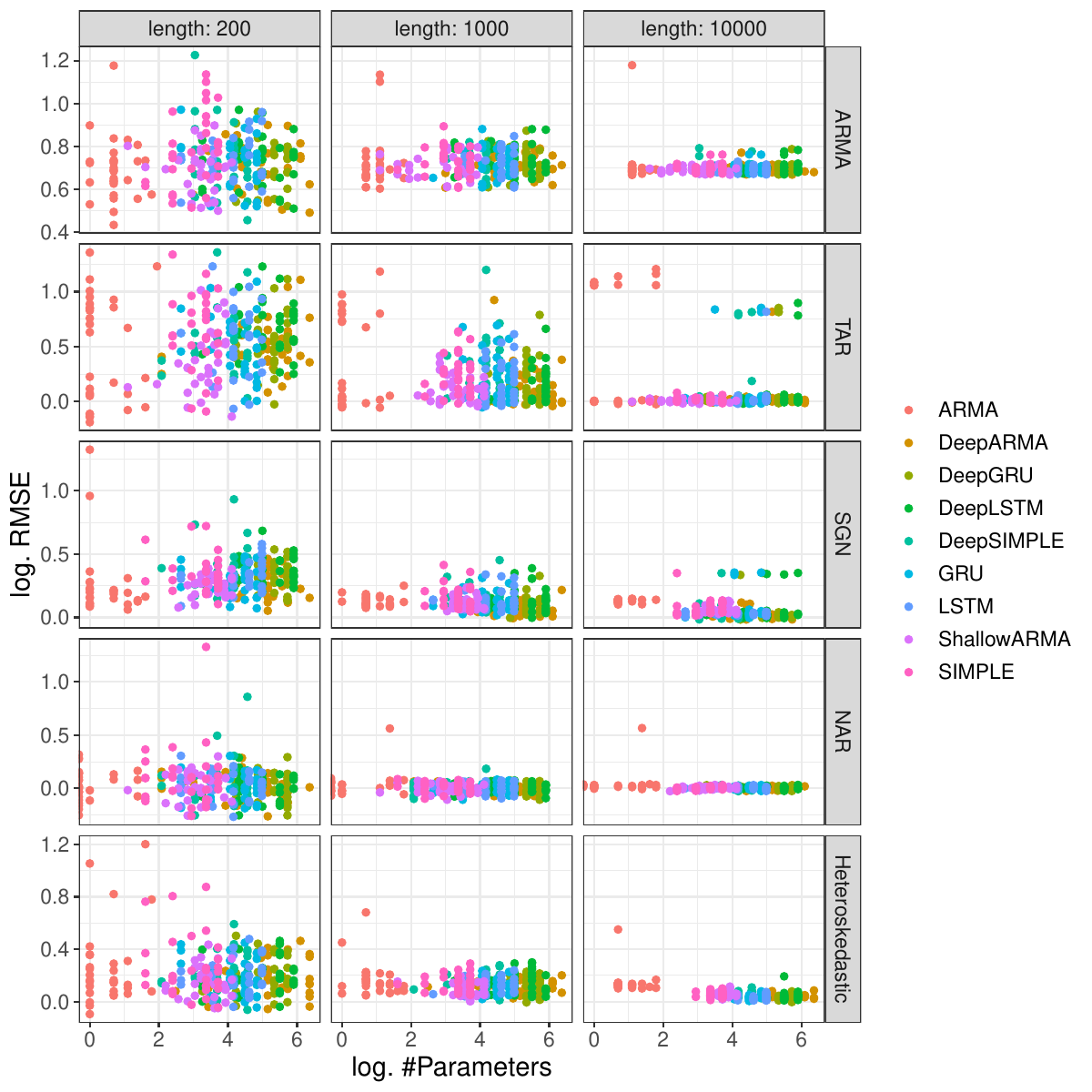}
    \caption{Logarithmic RMSE values (y-axis) for different models (colors) and their respective parameter numbers (x-axis) separated by the time series length (columns) and data sets (rows).}
    \label{fig:param_len}
\end{figure}

In Figure~\ref{fig:param_len} we compare the logarithmic RMSE values of
different models to the (logarithmic) number of parameters in the case of a
one-step forecasting horizon. Whereas performance values are very similar for
all models for the linear time series (ARMA, TAR), the ShallowARMA model yields
better results than the SIMPLE, GRU, and LSTM models while having fewer
parameters compared to the latter two architectures. The DeepARMA model often
yields a similar or larger number of parameters compared to the other deep
architectures, while also yielding smaller RMSE values, in particular for
non-linear time series (NAR, SGN).

\begin{figure}
    \centering
    \includegraphics[width=0.6\textwidth]{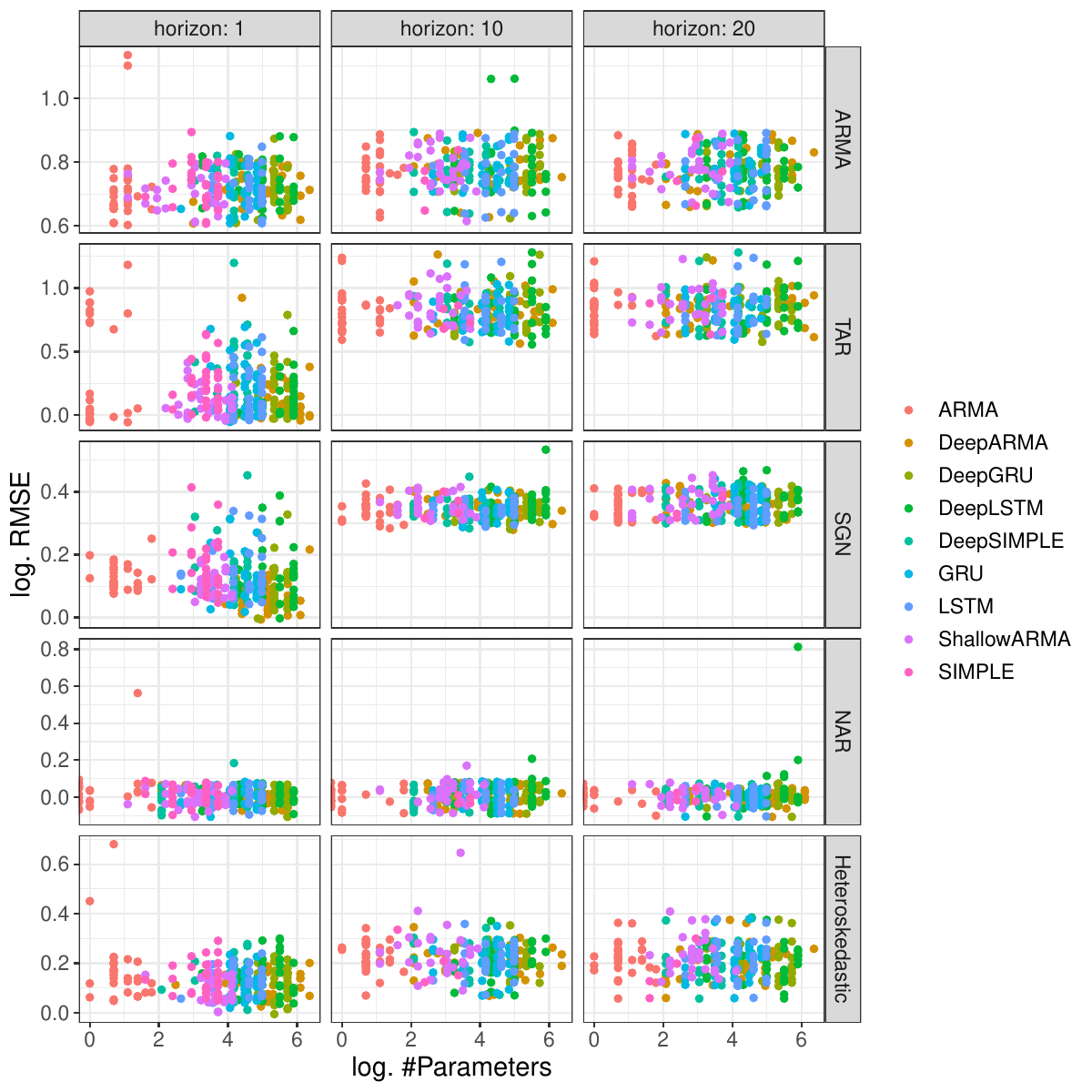}
    \caption{Logarithmic RMSE values (y-axis) for different models (colors) and their respective parameter numbers (x-axis) separated by the forecasting horizon (columns) and data sets (rows).}
    \label{fig:param_hor}
\end{figure}

\begin{figure}
    \centering
    \includegraphics[width=0.6\textwidth]{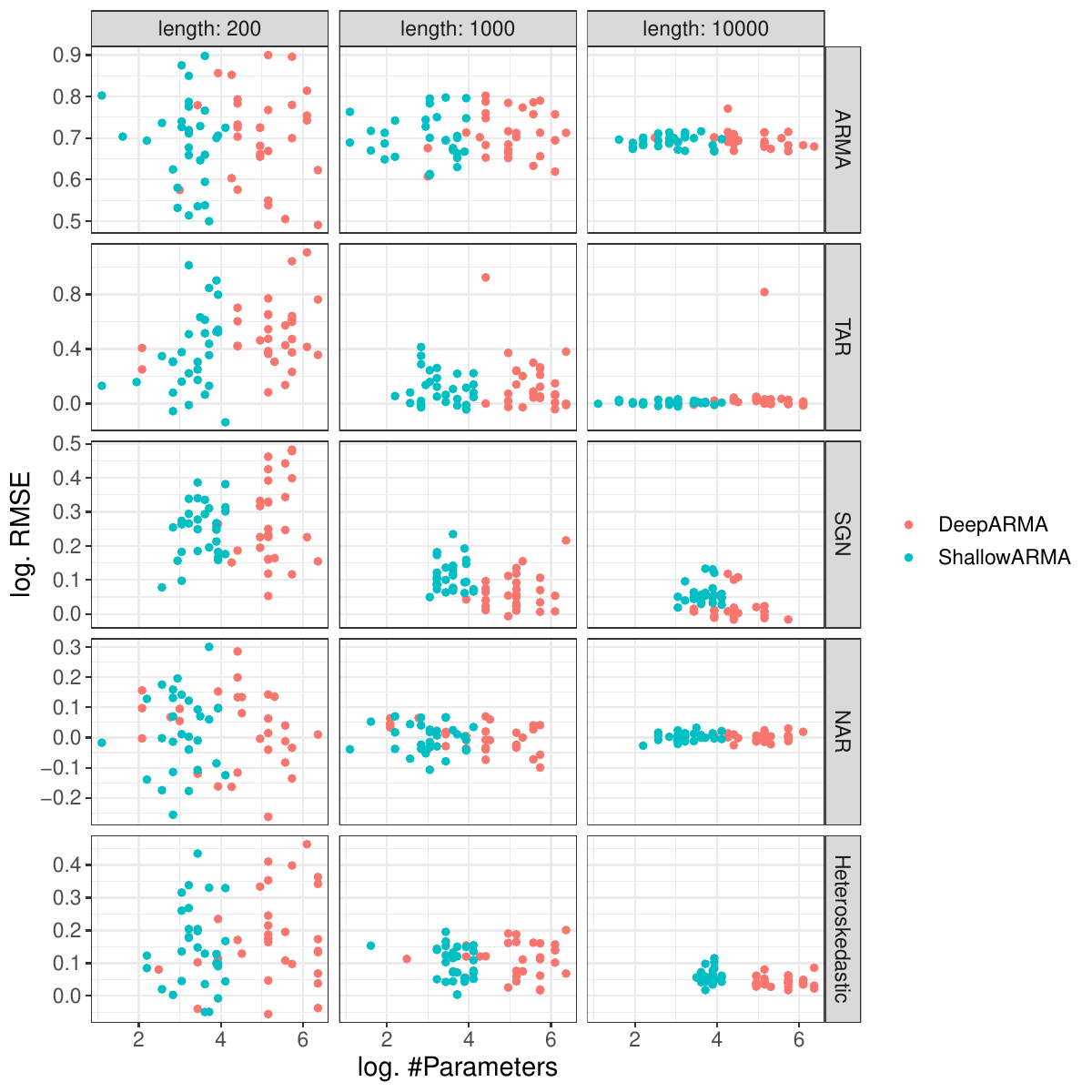}
    \caption{Logarithmic RMSE values (y-axis) for the two different ARMA models (colors) and their respective parameter numbers (x-axis) separated by the time series length (columns) and data sets (rows).}
    \label{fig:param_arma}
\end{figure}

A similar result can be observed for different forecasting horizons
(Figure~\ref{fig:param_hor}). When comparing only the two ARMA-variants
(Figure~\ref{fig:param_arma}), the improvement using a deep instead of a shallow
ARMA becomes apparent, but -- as expected -- only on the non-linear data sets
(TAR, SGN) and larger time series (1000, 10000).

\begin{figure}
    \centering
    \includegraphics[width=0.9\textwidth]{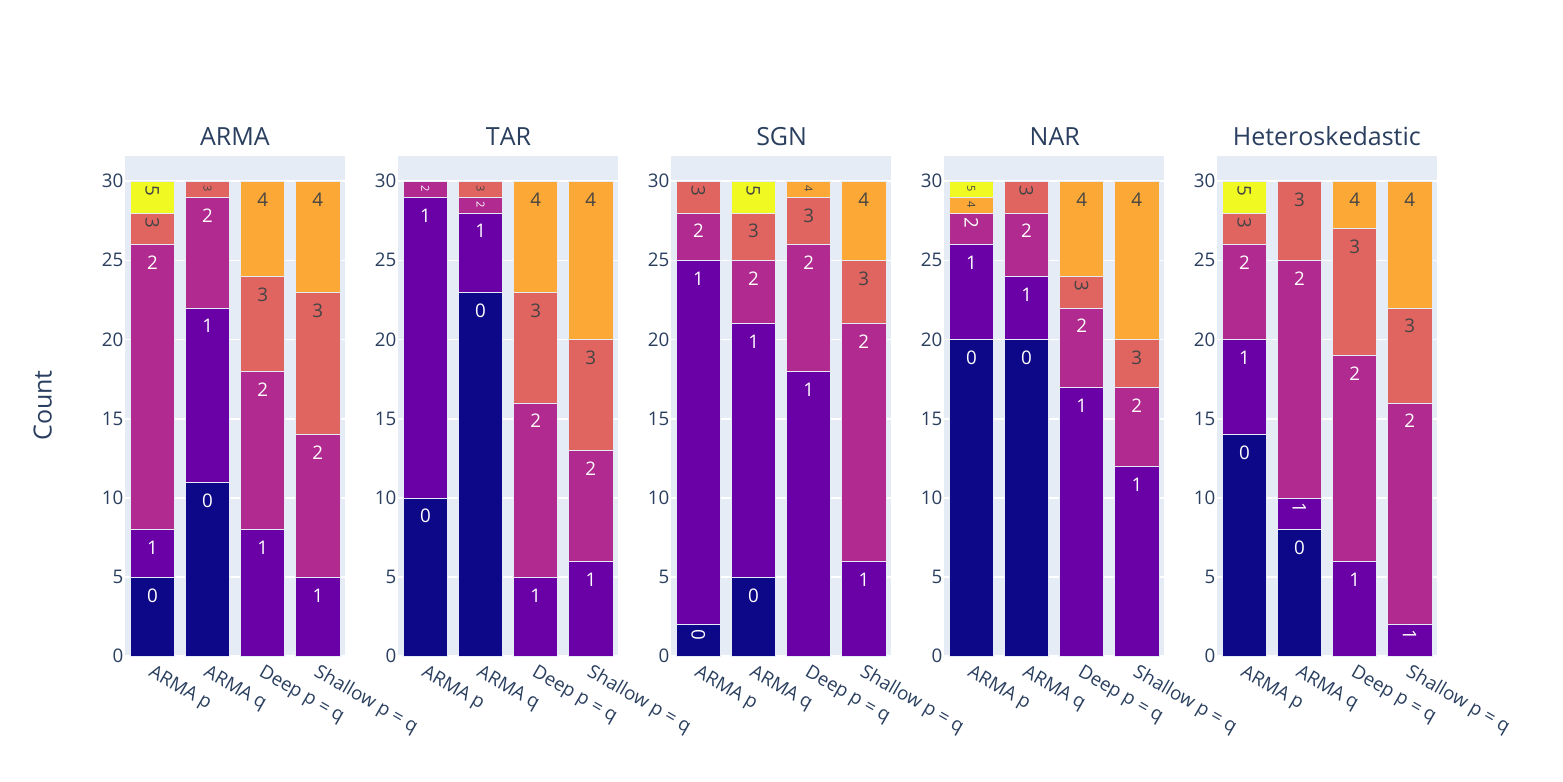}
    \caption{Aggregated lag choices (number of counts on the y-axis for $p$ and $q$ for the classical ARMA model, as well as the ShallowARMA and DeepARMA for each of the 5 data sets (different plots)).}
    \label{fig:pq_choices}
\end{figure}

We now further investigate the chosen number of lags $p$ and $q$ in the
hyperparameter optimization routine. Figure~\ref{fig:pq_choices} summarizes the
result of this analysis by plotting the number of runs in which a given value
for $p$ or $q$ was chosen. As we have the simplifying assumption of only
considering $p=q$ in the ShallowARMA and DeepARMA models, only one bar is given
for each model. We find that the classical ARMA model chooses the correct value
of $p=2$ in the majority of cases. We further find that the DeepARMA model tends
to use a lower number of lags compared to the ShallowARMA model, indicating that
the second layer facilitates more complex model spaces that are otherwise
captured by a longer lag structure.

\paragraph{Influence of the non-linearity.}
To assess the influence of the non-linear activation, we check how often a
purely linear activation was chosen by the hyperparameter optimization for the
ShallowARMA model in the univariate simulations. We find that the hyperparameter
optimization opted to use non-linear activations in almost all cases, only using
linear models 3 times for NAR, 1 time for Heteroskedastic, and 0 times for TAR
and SGN. For the ARMA time series, however, a purely linear activation was
selected more often, in 9 out of 30 cases. This indicates that the non-linearity
contributes to the performance improvement of the cell. By contrast, we did not
find substantial differences based on the number of units, as shown in
Figures~\ref{fig:param_len}-\ref{fig:param_arma}. Note that while these figures
show the number of parameters, this implicitly groups them by the number of
units.

\subsection{Additional simulation and benchmark results} \label{app:adben}

In the following, we provide additional results on numerical experiments by
including comparisons based on the mean absolute error (MAE).

\paragraph{Results.} The results for simulated data suggest that either the
ShallowARMA or DeepARMA cell perform best in most cases while on par with the
GRU cell for the ARMA and NAR dataset. For the simulated multivariate time
series, none of the existing neural methods outperforms the ARMA cells. For the
time series benchmark datasets, the ARMA approaches outperform all other RNN
approaches on M4 and Traffic. On the Electricity dataset the ARMA cells remain
competitive for the univariate case, but yield larger RMSE values compared to
GRU and Simple in the multivariate setting.

Overall the rankings of methods do not change notably when using the MAE instead
of the RMSE as comparison measure.

\begin{table}[t]
    \caption{Comparisons of different methods (rows) and different data
        generating processes (columns) for univariate time series using the average
        MAE $\pm$ the standard deviation of 10 independent runs. The best performing
        method is highlighted in bold, the second-best in italics.}
    \label{simulation_1_MAE}
    \begin{center}
        \begin{small}
            \begin{sc}
                \begin{tabular}{llllll}
                    \toprule
                    {}                   & ARMA                    & TAR
                                         & SGN                     & NAR           & Heteroskedastic \\
                    model                &                         &
                                         &                         &               &                 \\
                    \midrule
                    $\text{ARMA}$        & 1.62$\pm$0.29           &
                    2.39$\pm$2.66        & 1.98$\pm$2.35           & 2.28$\pm$4.42
                                         & 0.95$\pm$0.11                                             \\
                    $\text{ShallowARMA}$ & \bfseries 1.55$\pm$0.07 & \bfseries
                    0.84$\pm$0.06        & 0.88$\pm$0.06           & 0.81$\pm$0.04 & \bfseries
                    0.88$\pm$0.05                                                                    \\
                    $\text{DeepARMA}$    & 1.57$\pm$0.08           &
                    0.96$\pm$0.38        & \bfseries 0.84$\pm$0.05 &
                    0.81$\pm$0.04        & \itshape 0.88$\pm$0.05                                    \\
                    $\text{LSTM}$        & 1.57$\pm$0.07           &
                    1.08$\pm$0.36        & 0.96$\pm$0.15           & \itshape
                    0.81$\pm$0.04        & 0.92$\pm$0.08                                             \\
                    $\text{DeepLSTM}$    & 1.60$\pm$0.08           &
                    1.12$\pm$0.42        & 0.94$\pm$0.12           & 0.81$\pm$0.04
                                         & 0.92$\pm$0.07                                             \\
                    $\text{GRU}$         & \itshape 1.56$\pm$0.09  &
                    0.98$\pm$0.22        & 0.88$\pm$0.07           & \bfseries
                    0.81$\pm$0.04        & 0.90$\pm$0.06                                             \\
                    $\text{DeepGRU}$     & 1.58$\pm$0.07           & \itshape
                    0.95$\pm$0.26        & \itshape 0.87$\pm$0.10  & 0.81$\pm$0.04 &
                    0.90$\pm$0.05                                                                    \\
                    $\text{Simple}$      & 1.58$\pm$0.08           &
                    0.99$\pm$0.22        & 0.91$\pm$0.08           & 0.83$\pm$0.04
                                         & 0.90$\pm$0.06                                             \\
                    $\text{DeepSimple}$  & 1.60$\pm$0.08           &
                    1.15$\pm$0.46        & 0.92$\pm$0.08           & 0.82$\pm$0.04
                                         & 0.92$\pm$0.08                                             \\
                    \bottomrule
                \end{tabular}
            \end{sc}
        \end{small}
    \end{center}
    \vskip -0.1in
\end{table}

\begin{table}[t]
    \caption{Comparisons of different methods (rows) and different data
        generating processes (columns) for multivariate time series using the
        average MAE $\pm$ the standard deviation of 10 independent runs. The best
        performing method is highlighted in bold, the second-best in italics.}
    \label{simulation_2_MAE}
    \begin{center}
        \begin{small}
            \begin{sc}
                \begin{tabular}{llll}
                    \toprule
                    {}                   & VARMA                   & EXP
                                         & SQ                                  \\
                    model                &                         &
                                         &                                     \\
                    \midrule
                    $\text{VARMA}$       & \bfseries 0.80$\pm$0.02 &
                    1.61$\pm$0.13        & 1.22$\pm$0.07                       \\
                    $\text{ShallowARMA}$ & \itshape 0.80$\pm$0.03  & \itshape
                    1.48$\pm$0.13        & \bfseries 1.19$\pm$0.06             \\
                    $\text{DeepARMA}$    & 0.81$\pm$0.03           & \bfseries
                    1.48$\pm$0.13        & \itshape 1.20$\pm$0.06              \\
                    $\text{LSTM}$        & 0.81$\pm$0.03           &
                    1.55$\pm$0.13        & 1.25$\pm$0.09                       \\
                    $\text{DeepLSTM}$    & 0.82$\pm$0.03           &
                    1.63$\pm$0.16        & 1.29$\pm$0.10                       \\
                    $\text{GRU}$         & 0.81$\pm$0.03           &
                    1.53$\pm$0.14        & 1.23$\pm$0.07                       \\
                    $\text{DeepGRU}$     & 0.81$\pm$0.03           &
                    1.54$\pm$0.13        & 1.23$\pm$0.09                       \\
                    $\text{SIMPLE}$      & 0.82$\pm$0.02           &
                    1.56$\pm$0.15        & 1.23$\pm$0.07                       \\
                    $\text{DeepSimple}$  & 0.82$\pm$0.03           &
                    1.56$\pm$0.17        & 1.25$\pm$0.08                       \\
                    \bottomrule
                \end{tabular}
            \end{sc}
        \end{small}
    \end{center}
    \vskip -0.1in
\end{table}

\begin{table}[t]
    \caption{Comparison of different univariate and multivariate forecasting approaches (rows) for different datasets (columns) based on the average MAE $\pm$ the standard deviation of 10 independent runs. The best performing method is highlighted in bold, the second-best in italics.}
    \label{benchmark_MAE}
    \begin{center}
        \begin{small}
            \begin{sc}
                \begin{tabular}{clllll}
                    {}            &                         & M4                      & traffic
                                  & electricity                                                   \\
                    \midrule
                    \parbox[t]{2mm}{\multirow{9}{*}{\rotatebox[origin=c]{90}{univ.\,\,}}}
                                  & $\text{ARMA}$           & \bfseries 0.82$\pm$0.00 &
                    0.48$\pm$0.00 & 0.77$\pm$0.00                                                 \\
                                  & $\text{ShallowARMA}$    & \itshape 0.83$\pm$0.00  &
                    0.48$\pm$0.00 & 0.74$\pm$0.01                                                 \\
                                  & $\text{DeepARMA}$       & 0.84$\pm$0.01           & \bfseries
                    0.45$\pm$0.01 & \itshape 0.70$\pm$0.02                                        \\
                                  & $\text{LSTM}$           & 0.88$\pm$0.03           & \itshape
                    0.45$\pm$0.00 & 0.72$\pm$0.03                                                 \\
                                  & $\text{DeepLSTM}$       & 0.98$\pm$0.10           &
                    0.45$\pm$0.00 & \bfseries 0.69$\pm$0.02                                       \\
                                  & $\text{GRU}$            & 0.86$\pm$0.02           &
                    0.45$\pm$0.01 & 0.70$\pm$0.02                                                 \\
                                  & $\text{DeepGRU}$        & 0.86$\pm$0.02           &
                    0.45$\pm$0.01 & 0.70$\pm$0.01                                                 \\
                                  & $\text{Simple}$         & 0.91$\pm$0.06           &
                    0.46$\pm$0.00 & 0.71$\pm$0.01                                                 \\
                                  & $\text{DeepSimple}$     & 0.93$\pm$0.07           &
                    0.46$\pm$0.01 & 0.70$\pm$0.01                                                 \\
                    \toprule
                    \midrule
                    \parbox[t]{2mm}{\multirow{9}{*}{\rotatebox[origin=c]{90}{multiv.\,\,}}}
                                  & $\text{ARMA}$           & \itshape 0.82$\pm$0.00  &
                    0.48$\pm$0.00 & 0.77$\pm$0.00                                                 \\
                                  & $\text{ShallowARMA}$    & \bfseries 0.82$\pm$0.01 &
                    0.53$\pm$0.01 & 0.78$\pm$0.01                                                 \\
                                  & $\text{DeepARMA}$       & 0.83$\pm$0.00           &
                    0.53$\pm$0.02 & 0.77$\pm$0.03                                                 \\
                                  & $\text{LSTM}$           & 0.97$\pm$0.04           &
                    0.49$\pm$0.01 & 0.93$\pm$0.20                                                 \\
                                  & $\text{DeepLSTM}$       & 1.06$\pm$0.08           &
                    0.49$\pm$0.02 & 0.74$\pm$0.03                                                 \\
                                  & $\text{GRU}$            & 0.97$\pm$0.09           &
                    0.48$\pm$0.00 & 0.73$\pm$0.02                                                 \\
                                  & $\text{DeepGRU}$        & 0.97$\pm$0.06           &
                    0.48$\pm$0.01 & \itshape 0.72$\pm$0.02                                        \\
                                  & $\text{Simple}$         & 0.99$\pm$0.03           & \bfseries
                    0.47$\pm$0.01 & 0.73$\pm$0.02                                                 \\
                                  & $\text{DeepSimple}$     & 1.01$\pm$0.05           & \itshape
                    0.47$\pm$0.01 & \bfseries 0.70$\pm$0.01                                       \\
                    \bottomrule
                \end{tabular}
            \end{sc}
        \end{small}
    \end{center}
    \vskip -0.1in
\end{table}

\end{document}